\documentclass[11pt,a4paper]{article}

\usepackage[a4paper,top=0.8in,bottom=0.8in,left=0.8in,right=0.8in]{geometry}

\usepackage[utf8]{inputenc}

\usepackage[x11names]{xcolor}
\usepackage{soulpos}

\usepackage{amsmath, amssymb, mathtools, bm, bbm, relsize}

\usepackage{booktabs, multirow, tabularx, ragged2e, colortbl, makecell}

\usepackage{graphicx, wrapfig, titlepic}
\usepackage{tikz}
\usetikzlibrary{
  positioning,
  shapes.geometric,
  shapes.misc,
  shapes.symbols,
  arrows.meta,
  fit,
  calc
}

\usepackage{enumitem}

\usepackage{listings}
\usepackage{fancyvrb}

\usepackage[most]{tcolorbox}

\usepackage[font=small]{caption} 

\usepackage{csquotes}

\usepackage{setspace}
\usepackage{xspace}

\usepackage{XCharter}
\usepackage[scaled=1.1]{zlmtt}
\usepackage[xcharter,bigdelims,vvarbb]{newtxmath}

\usepackage{footmisc}
\usepackage{marginnote}

\usepackage{fancyhdr}
\usepackage{titlesec, titling}
\usepackage{tocloft, etoc}
\usepackage{lscape}

\usepackage{lipsum}

\usepackage{kotex}

\usepackage{xparse}

\usepackage[
  backend=biber,
  style=nature,
  sorting=none,
  maxbibnames=3,
  natbib=true,
  backref=true,
]{biblatex}

\usepackage[
  pagebackref=false,
  breaklinks=false,
  colorlinks=true,
  bookmarks=true,
  citecolor=Blue3,
  urlcolor=Blue3,
  linkcolor=Blue3,
]{hyperref}
\usepackage[noabbrev,capitalize]{cleveref}

\let\oldsection\section
\makeatletter
\newcounter{@secnumdepth}
\RenewDocumentCommand{\section}{s o m}{%
  \IfBooleanTF{#1}
    {\setcounter{@secnumdepth}{\value{secnumdepth}}
     \setcounter{secnumdepth}{0}
     \oldsection{#3}
     \setcounter{secnumdepth}{\value{@secnumdepth}}}
    {\IfValueTF{#2}
       {\oldsection[#2]{#3}}
       {\oldsection{#3}}}
}
\makeatother

\let\oldsubsection\subsection
\makeatletter
\newcounter{@subsecnumdepth}
\RenewDocumentCommand{\subsection}{s o m}{%
  \IfBooleanTF{#1}
    {\setcounter{@subsecnumdepth}{\value{subsecnumdepth}}
     \setcounter{subsecnumdepth}{0}
     \oldsubsection{#3}
     \setcounter{subsecnumdepth}{\value{@subsecnumdepth}}}
    {\IfValueTF{#2}
       {\oldsubsection[#2]{#3}}
       {\oldsubsection{#3}}}
}
\makeatother

\fancypagestyle{maincontent}{
  \fancyhf{}
  \fancyhead[L]{\small\textit{Spacer}: Towards Engineered Scientific Inspiration}
  \fancyfoot[C]{\small\thepage}
}

\titleformat{\section}
  {\normalfont\Large\bfseries}{\thesection.}{1em}{}
\titlespacing*{\section}{0pt}{*3}{*1.5}

\titleformat{\subsection}
  {\normalfont\large\bfseries}{\thesubsection}{1em}{}
\titlespacing*{\subsection}{0pt}{*2}{*1}

\definecolor{blueheader}{rgb}{0.2078, 0.3608, 0.6314}
\definecolor{bluebg}{rgb}{0.9490, 0.9647, 0.9804}
\definecolor{blackheader}{rgb}{0.2588, 0.2588, 0.2588}
\definecolor{blackbg}{rgb}{0.9804, 0.9804, 0.9804}

\begin{document}

\title{\vspace{.2cm}\textbf{\textit{Spacer}: Towards Engineered Scientific Inspiration}}
\author{\small Asteromorph\thanks{See \hyperref[sec:contributions]{Contributions and Acknowledgments}.}}
\date{}
\maketitle
\vspace{-1.5cm}

\fancyfoot[C]{}
\thispagestyle{fancy}


\linespread{1.3}
\renewcommand{\abstractname}{}
\begin{abstract}
    \footnotesize
    Recent advances in LLMs have made automated scientific research the next frontline in the path to artificial superintelligence.
    However, these systems are bound either to tasks of narrow scope or the limited creative capabilities of LLMs.
    We propose \textit{Spacer}, a scientific discovery system that develops creative and factually grounded concepts without external intervention.
    \textit{Spacer} attempts to achieve this via `deliberate decontextualization,' an approach that disassembles information into atomic units---keywords---and draws creativity from unexplored connections between them.
    \textit{Spacer} consists of (i) \textsc{Nuri}, an inspiration engine that builds keyword sets, and (ii) the Manifesting Pipeline that refines these sets into elaborate scientific statements.
    \textsc{Nuri} extracts novel, high-potential keyword sets from a keyword graph built with 180,000 academic publications in biological fields.
    The Manifesting Pipeline finds links between keywords, analyzes their logical structure, validates their plausibility, and ultimately drafts original scientific concepts.
    According to our experiments, the evaluation metric of \textsc{Nuri} accurately classifies high-impact publications with an AUROC score of 0.737.
    Our Manifesting Pipeline also successfully reconstructs core concepts from the latest top-journal articles solely from their keyword sets.
    An LLM-based scoring system estimates that this reconstruction was sound for over 85\% of the cases.
    Finally, our embedding space analysis shows that outputs from \textit{Spacer} are significantly more similar to leading publications compared with those from SOTA LLMs.
\end{abstract}

\begin{figure}[h!]
    \centering
    \includegraphics[width=0.85\textwidth]{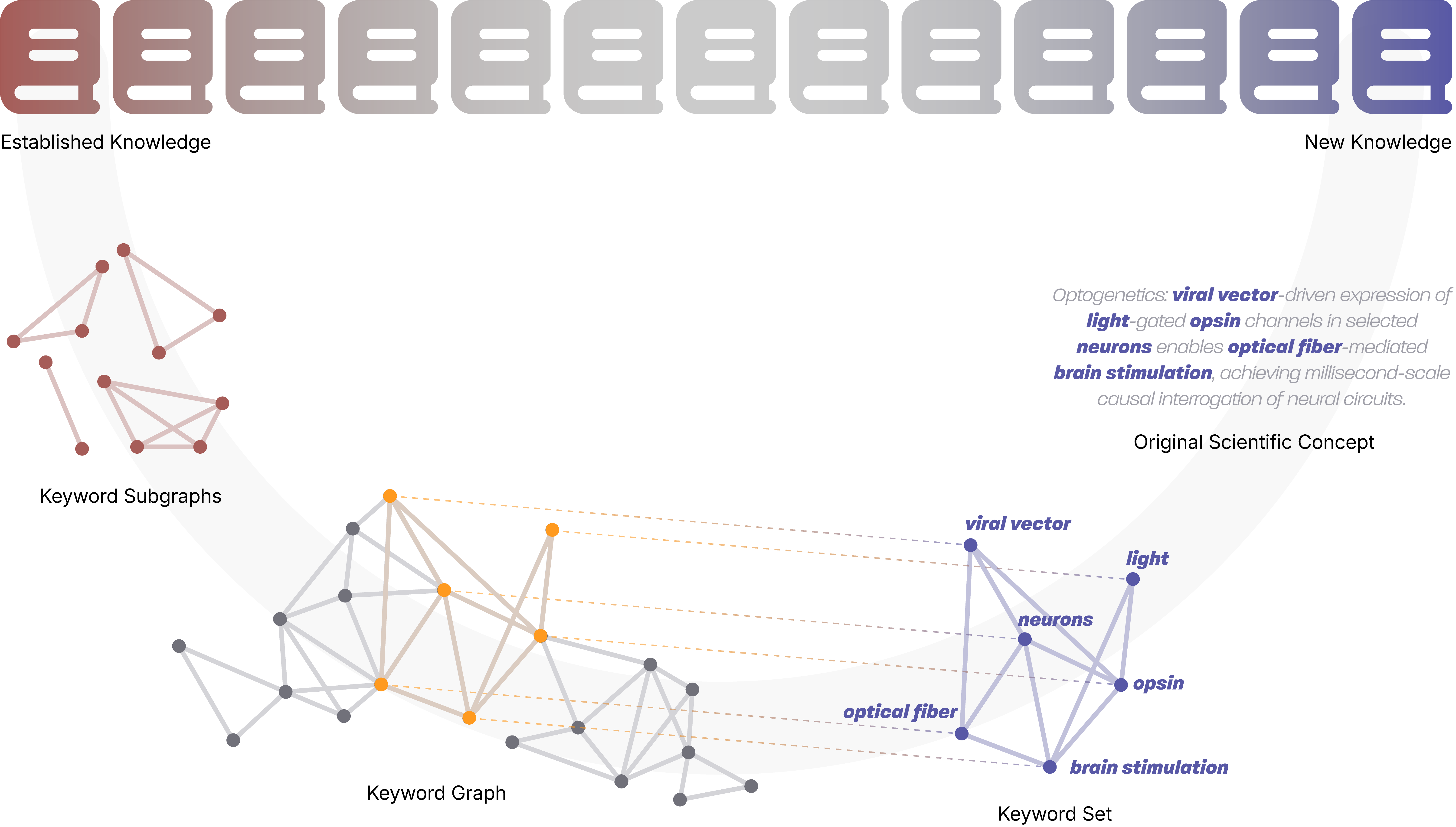}
    \vspace{0.3cm}
    \caption{Schematic of \textit{Spacer}'s approach to engineered scientific inspiration.}
    \label{fig:approach_diagram}
\end{figure}

\newpage

\fancyfoot[C]{\thepage}
{\hypersetup{linkcolor=black}
    \tableofcontents
}
\thispagestyle{maincontent}

\newpage
\pagestyle{maincontent}

\section{Introduction}

Throughout history, scientific breakthroughs have emerged from the conjunction of seemingly disparate fields of knowledge \cite{shibata2008-fa, uzzi2013-ga, youn2015-id, fortunato2018-mr}.
Optogenetics \cite{boyden2005-oe, deisseroth2011-mx, yizhar2011-pk} has revolutionized neuroscience by introducing light-mediated modulation in neural cells.; CRISPR-Cas9 \cite{jinek2012-mw, qi2013-ea, doudna2014-fy} has changed the landscape of biological research by reinterpreting the bacterial immune system as a platform for genome editing.
Thomas Kuhn characterized such moments as `paradigm shifts,' arguing that these fundamental reorientations of scientific understanding cannot emerge through incremental progress.
Over the last few decades, academia has witnessed an unprecedented surge in the sheer volume of scholarly publications \cite{lewis2021unesco, nsb2023publications}; despite this, innovations on par with Kuhn's portrayal have been rare \cite{park2023-lg}.

Recently, large language models (LLMs) have garnered traction as potential galvanizers of creativity.
Supporting this sentiment is the notable performance of LLMs in various benchmarks that measure capabilities in science, programming, and reasoning \cite{rein2023gpqagraduatelevelgoogleproofqa, hendrycks2021measuringmassivemultitasklanguage, phan2025humanitysexam, jimenez2024swebenchlanguagemodelsresolve, chen2021evaluatinglargelanguagemodels}.
There have already been several attempts to capitalize on these advancements by creating agentic frameworks for scientific discovery \cite{gottweis2025aicoscientist, lu2024aiscientistfullyautomated, yamada2025aiscientistv2workshoplevelautomated, ghareeb2025robinmultiagentautomatingscientific, m_bran2024-ff, sim2024-vc, fehlis2025acceleratingdrugdiscoveryagentic}.
AlphaEvolve \cite{novikov2025alphaevolvecodingagentscientific} finds solutions for local optimization problems, and a multi-agent `Virtual Lab' \cite{swanson2025-rc} uncovers molecule-level candidates for SARS-CoV-2 nanobodies.

However, we remain skeptical about whether systems that rely solely on LLMs can initiate artificial paradigm shifts.
Transformer-based architectures optimize contextual coherence and penalize outputs that deviate from established patterns.
Skewed evaluation metrics and human feedback further reinforce this behavior, leading to a complementary systematic bias that favors soundness over novelty \cite{kirk2024understandingeffectsrlhfllm,sharma2025understandingsycophancylanguagemodels}.
The problem is that while building upon patterns, attempts at creative thinking easily degrade into regressions.
For instance, the term `CRISPR-Cas9' often recurs when prompted to generate research ideas, due to the overrepresentation of the CRISPR-Cas9 technology as novel research in training datasets.
As a result, LLM outputs tend to slant toward the precursory prompt context and the training data, implying that an automated ideation system powered by LLMs must overcome the limitations introduced by contextualization.

To this end, we decompose information into atomic units: keywords.
As keywords do not carry excess context, we may leverage the reasoning abilities of LLMs while circumventing the aforementioned limitations.
This decomposition also allows for the compositional construction of scientific concepts, where keywords act as versatile building blocks that comprise diverse combinations.
Known concepts correspond to specific combinations of keywords that come together into a graph: one that we believe spans humankind's ever-growing knowledge base.
Furthermore, we assert that we can automate paradigm-shifting discoveries through the structural examination of this graph.

As such, we propose \textit{Spacer}, a two-step scientific discovery system designed to foster probable but unexplored connections between seemingly unrelated concepts.
\textsc{Nuri}, an inspiration engine, extracts potent sets of keywords from the global keyword graph.
The Manifesting Pipeline then substantiates each set into a concrete scientific statement that potentially expands the boundaries of science.

\clearpage
\section{Spacer\label{sec:spacer}}

\textit{Spacer} is a \textbf{scientific discovery system}.
The goal of \textit{Spacer} is to generate original scientific concepts distinct from those seen in its training datasets or search results, while complying with academic standards.
Whereas next-token prediction models have been unsuccessful in creating original scientific concepts, \textit{Spacer} overcomes their limitations by operating on a multi-stage pipeline with both LLM and non-LLM components.


\subsection{Overall Approach}

The core conceptual and technical backbone of \textit{Spacer} is the \textbf{deliberate decontextualization} of information.
We define deliberate decontextualization as the intentional removal of inferable contextual information: sentences, paragraphs, or any structured text that would encourage the language model to rely on familiar patterns or established knowledge.
We have approached the problem of developing \textit{Spacer} based on the decontextualization of information for the following two reasons.
First, significant scientific breakthroughs often emerge from unexpected connections between seemingly unrelated concepts, rather than from next-step extensions of existing knowledge.
Second, decontextualization enables a multi-LLM framework to generate contexts beyond its knowledge span.

LLMs' next-token prediction engines are antithetical to the concept of decontextualization.
As mentioned above, the outputs of LLMs are constructed upon heavy contextual bias when given a specific task.
While excelling in logical writing and elaboration, they cannot make new conceptual connections necessary for scientific inspiration.
Therefore, \textit{Spacer} conceives scientific inspiration before introducing LLMs.
A non-LLM component first generates novel connections between concepts from decontextualized information, then passes them to LLMs for construction into scientific concepts.
Completely separating the ideation from LLMs, the workflow preserves the emergent properties while developing scientific significance using multi-agent LLM frameworks.

The atomic unit of decontextualization in \textit{Spacer} is a word.
That is, we seek original knowledge in the combinations of scientific terms.
Regarding our assertion that unexpected conceptual connections can lead to scientific breakthroughs, a set of scientific terms is a fine starting point for ideas to accumulate.
From the perspective of format, words---the results of decontextualization---are the optimal point for LLMs to operate on.
A token does not preserve the original meaning of words; larger units, such as phrases or sentences, carry contextual bias.
Therefore, a set of words is a format that manifests scientific inspiration while allowing LLMs to understand and develop meaning without contextual bias.

To summarize, we deliberately decontextualize scientific information to free LLMs from contextual dependencies.
We adopt an hybrid approach where LLMs only build upon a given set of scientific words that already contains emergent knowledge.
Such an approach enables the system to access novel scientific discoveries while also maintaining logical soundness and scientific compatibility.
This meet-in-the-middle approach is implemented in \textit{Spacer} as three consecutive stages: word selection, word combination\footnote{
    One might argue that involving LLMs in the word combination stage is contradictory to the principle of deliberate decontextualization.
    However, since \textit{Spacer} devotes inspiration to the word selection stage, there are no `creative' tasks involved in word combination.
}, and agentic reinforcement.
Scientific knowledge is decontextualized into words, then selected to generate new connections and findings.

All emergent characteristics of \textit{Spacer} originate from the word selection stage.
The following steps involve LLMs to combine words into sentences that reveal the idea and reinforce it with rationales and logical refinement.
The specific architecture and design of the frameworks are discussed in depth in the following section.


\subsection{Architecture}

We first define the key terms used throughout this report, each corresponding to a different level of scientific discovery.
\textbf{Keywords} are individual decontextualized words that carry specific meaning.
Biological entities, scientific techniques, or established concepts may all serve as keywords.
\textbf{Theses} are individual paragraphs aiming to capture a researcher's initiative.
These may resemble abstracts but omit resultant details such as findings, data, expected impacts, or speculative outcomes.
Finally, when Theses are organized into structured formats, they form \textbf{Statements}: 1--2-sentence-length concepts accompanied by a supporting set of rationales.

\begin{figure}[htb]
    \centering
    \includegraphics[width=.95\textwidth]{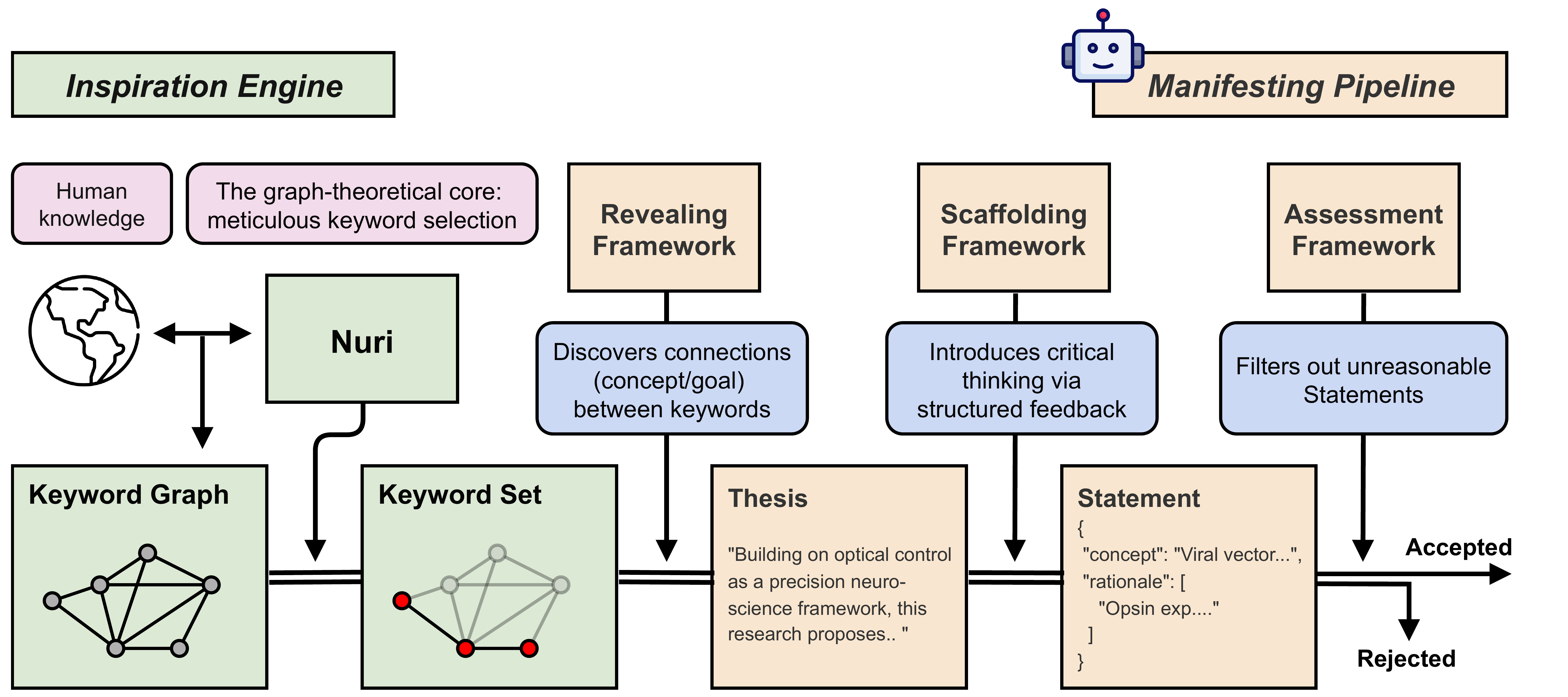}
    \caption{Architecture of \textit{Spacer}.}
    \label{fig:architecture}
\end{figure}

\textit{Spacer} consists of four primary components organized in a sequential pipeline.
\textbf{\textsc{Nuri}} builds a set of keywords that involve semantical richness while minimizing contextual dependencies.
The \textbf{Revealing Framework} transforms these keywords into Theses, which are then elevated to Statements by the \textbf{Scaffolding Framework}.
Lastly, the \textbf{Assessment Framework} evaluates Statements to accept or reject them based on their overall plausibility.
This design allows \textit{Spacer} to separate creative generation from critical evaluation---\textsc{Nuri} is devoted to addressing `creativity,' while the others focus on materializing a Statement without subverting the original concept.
This architecture is illustrated in Figure~\ref{fig:architecture}.

The continuing sections establish each of the four components in detail.


\newcommand{\kw}{\mathcal{K}}
\newcommand{\fwci}{\operatorname{FWCI}}

\subsubsection*{Nuri\label{subsubsec:nuri}}

\textsc{Nuri}\footnote{The name Nuri originates from the Korean word for `world'.} is a graph-based Keyword Set extraction algorithm.
We designed \textsc{Nuri} to select Keyword Sets, as connections between keywords reveal information not evident in individual keywords.
We believe that complex scientific ideas---such as the motivations behind existing research papers---can be represented in the form of these connections.
Accordingly, we designed \textsc{Nuri} to collect Keyword Sets that are likely to harness novel research directions that lead to impactful research papers.

For published papers, we employed the Field-Weighted Citation Impact (FWCI) as a primary metric for impactfulness.
FWCI compares the number of citations a paper receives against the average number of citations in the same field and year.
For example, an FWCI value greater than $1$ indicates that the paper has been cited more frequently than average.
As citation practices differ across fields, FWCI offers a compensation that allows for a fair comparison of scientific impact.
We had also considered using the Relative Citation Ratio (RCR), another index that rates impactful papers based on citation rate counts \cite{hutchins029629}.
However, given that these two indices are mostly interchangeable with each other \cite{purkayastha2019635}, we chose FWCI over RCR for its abundance in precomputed values.

To be precise, \textsc{Nuri} makes a graph with set of papers, and an evaluation function which can determine whether a Keyword Set would be impactful.
For a given set of papers $P$, let $\mathcal{K}(p)$ be the Keyword Set of a paper $p \in P$.
We define the undirected and weighted graph $G(P)$ as
\begin{align}
    V(G(P)) & = \bigcup_{p\in P} \mathcal{K}(p), \\
    w(u, v) & = \sum_{\substack{p \in P          \\ u, v \in \kw(p)}} \frac{\log_2(\fwci(p) + 1)}{|\kw(p)| - 1}.
\end{align}

The weight of each edge represents the joint academic impact of the two keywords as a sum of normalized logarithmic FWCI values of the papers containing both vertices.
The evaluation function $f_P$ takes a Keyword Set $K \subset V(G(P))$ and outputs a normalized score $s \in [0, 1]$, representing the potential impact of the input keywords.
\textsc{Nuri} uses $f_P$ and applies various heuristics to create its final output set.
We note that \textsc{Nuri} does not invoke any machine learning methods or LLMs in its process, nor does it require any input from the user.


\subsubsection*{The Revealing Framework\label{subsubsec:revealing}}

\begin{figure}[htb]
    \centering
    \includegraphics[width=0.7\textwidth]{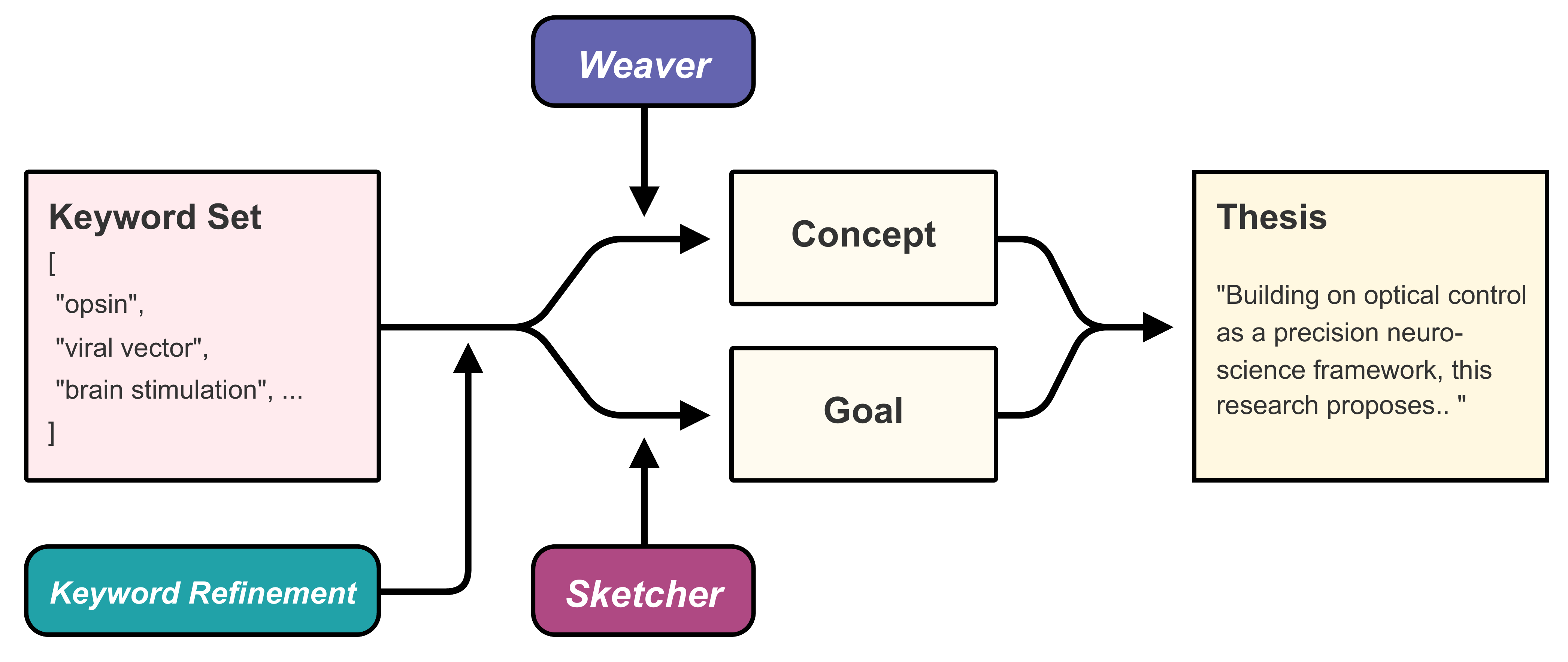}
    \caption{Schematic of the Revealing Framework.}
    \label{fig:revealing_framework}
\end{figure}

The Revealing Framework takes a Keyword Set and finds a plausible interconnection between them, forming a Thesis.
The input keywords selected by the preceding engine---\textsc{Nuri}---have the potential to establish unexplored scientific propositions.
When applied to such keywords, the Revealing Framework realizes the latent concept in the connections and formulates it into communicable natural language.

The primary component of the Revealing Framework is \textsc{Weaver}.
\textsc{Weaver} is an LLM trained on academic papers, fine-tuned to reconstruct research initiatives from segments of these papers.
Taking a Keyword Set from \textsc{Nuri}, \textsc{Weaver} materializes a sentence-length research concept from the knowledge concealed within the word set.
As shown in Section~\ref{subsec:paper_reconstruction}, \textsc{Weaver} yields logical, novel, and plausible research initiatives from well-structured input keywords.
To stabilize the quality of outputs from \textsc{Weaver}, we applied a keyword refinement engine to the original Keyword Set, vetting and replacing inadequate or unnecessary keywords before the set could enter \textsc{Weaver}.

However, \textsc{Weaver}'s outputs often lack a clear directive necessary for impactful research.
As such, the Revealing Framework utilizes another fine-tuned LLM named \textsc{Sketcher} in parallel to \textsc{Weaver}.
\textsc{Sketcher} is responsible for providing an overarching purpose to the resulting Thesis by generating a sentence-length research goal based on the given keywords.
While \textsc{Sketcher} is trained to suggest a broad and unbiased research goal, in application, it is supplied with the same refined Keyword Set as \textsc{Weaver} for scientific coherence.

Combining the research concept and the research goal, an untrained LLM generates a final paragraph-long Thesis.
The prompt used here was designed to avoid any bias towards specific academic fields or the research concept while retaining the insights and contributions originating from the preceding stages.
Putting everything together, the Revealing Framework reliably reveals new knowledge in the form of Theses from the Keyword Sets proposed by \textsc{Nuri}.


\subsubsection*{The Scaffolding Framework\label{subsubsec:scaffolding}}

\begin{figure}[htb]
    \centering
    \includegraphics[width=0.8\textwidth]{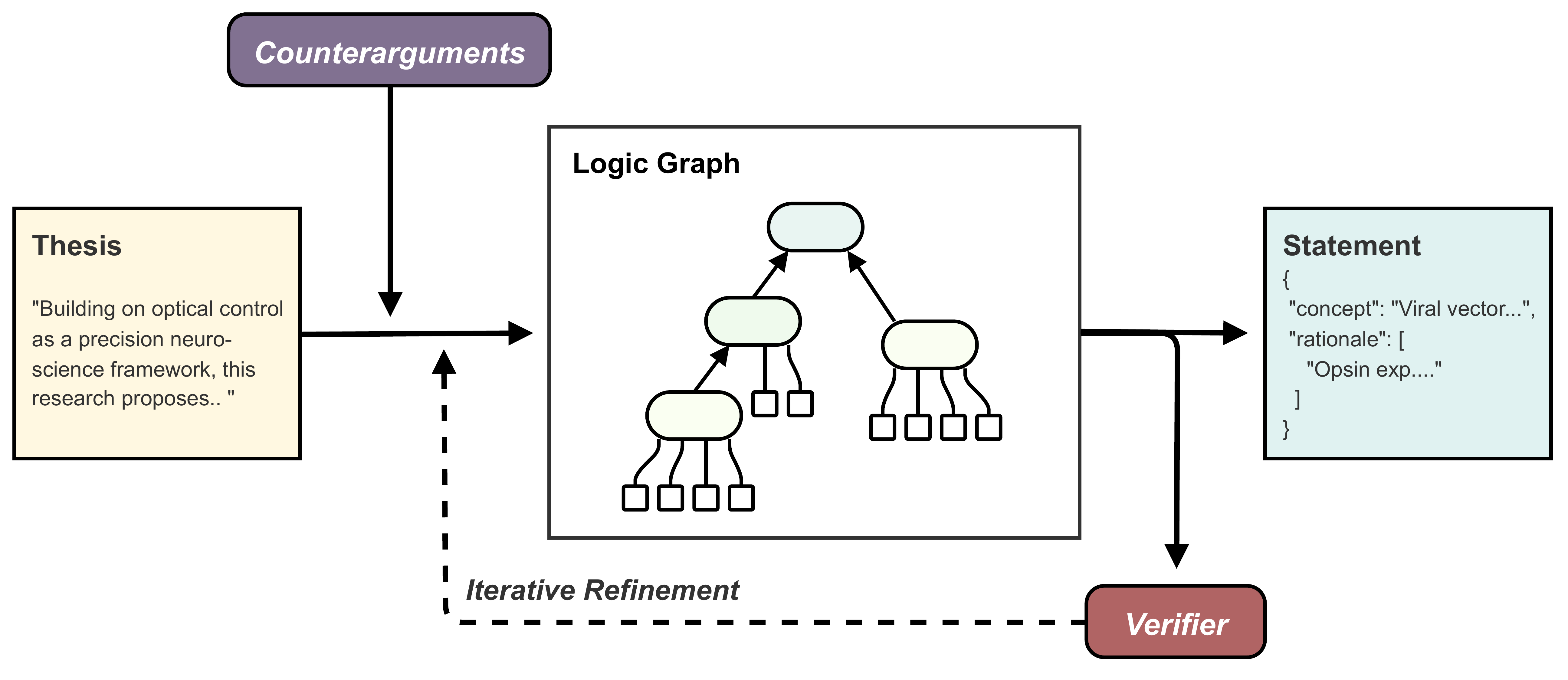}
    \caption{Schematic of the Scaffolding Framework.}
    \label{fig:scaffolding_framework}
\end{figure}

The Scaffolding Framework assembles Statements from their unstructured counterparts, notably supplementing them with validated evidence.
This framework employs logic graphs for its core mechanism.
Pieces of information---such as key concepts, supporting evidence, and intermediate conclusions---comprise typed vertices; relationships between these pieces form directed edges, constituting the logic graph.
A multiple-stage process built around this graph yields a Statement with a single concept and a set of rationales, where the concept is a 2--3-sentence-length summary of the Thesis, and each rationale is a description that validates the concept.

The need for a structured format emerges because LLMs yield suboptimal results when repeatedly processing context-heavy natural language.
While the Revealing Framework generates content rich in ideas, we observed that its irregular composition negatively impacts structural refinement and evaluation due to inherent biases.
When reading a paragraph, LLMs focus on connotations and nuances rather than the logical structure.
This phenomenon exacerbates in multi-agent feedback loops where irrelevant preferences amass and ultimately diverge from the constructive criticism we desire.
The Scaffolding Framework plays a crucial role in avoiding these shortcomings.

The first step in the Scaffolding Framework is the augmentation phase, where the Theses are challenged with counterarguments, each using peer-reviewed literature based on predefined criteria.
The Theses are then augmented to account for the identified points while preserving the original approach.

The following step is the graph iteration phase, where the paragraph is decomposed into a logic graph.
This phase iteratively reinforces the logic graph while verifying each revised graph against known facts.

When the graph iteration is deemed complete, the logic graph is reconfigured into a Statement in the final step.
The central vertices map to the concept, and the surrounding nodes translate into the rationales.
As such, the Scaffolding Framework rearranges the emergent idea from the Revealing Framework into Statements that are factual and logically sound.


\subsubsection*{The Assessment Framework}

The Assessment Framework is a system that evaluates the validity of Statements generated by the Scaffolding Framework.
Assessing scientific validity is inherently difficult; it depends on complex, context-specific factors, and rigid evaluation criteria often impose substantial bias, making consistent judgments elusive.
To mitigate this, the Assessment Framework is designed to evaluate Statements in two subsequent phases: exploratory analysis and specified inspection.

In the exploratory analysis phase, a reviewer LLM agent produces several critiques of a given Statement.
To avoid constraining the perspective of this agent, it is not provided with an explicit list of criteria.
Rather, the LLM agent is instructed to generate an unconstrained number of critiques together from a broader viewpoint, along with supporting rationales.

These reviews are carried forward to the specified inspection phase, where a meta-reviewer LLM agent evaluates them against predefined criteria, including practical feasibility, methodological rigor and fundamental scientific plausibility.
Each critique is independently assessed on how severely the flaw undermines the Statement's scientific integrity.
By steering the exploratory analysis clear of contextual biases, we can ensure that the evaluation correctly reflects the overall context and remains unbiased by specific criteria.

Following the process above, Statements that clear all initial critiques without substantial issues are determined to be scientifically sound and technologically feasible.
In human-expert evaluations, the Assessment Framework achieved an 88.2\% recall, highlighting its ability to detect latent flaws in superficially plausible yet scientifically unsound Statements.
In this way, it ensures that the Statements from the Scaffolding Framework remain not only innovative but also firmly grounded in scientific reality, thus enhancing the overall quality and reliability of the research outputs.

\clearpage
\section{Results}\label{sec:results}

In this section, we showcase the Statements generated by \textit{Spacer}.
Each output follows a structured format, consisting of the core concept, supporting rationales, and references.
For illustration, we present three examples with brief explanations and comments.
The explanations are based solely on the raw outputs with only minimal background information provided for clarity.

These examples demonstrate \textit{Spacer}'s capability for scientific discovery.
Through deliberate decontextualization and sophisticated reasoning, \textit{Spacer} synthesizes interdisciplinary concepts from a Keyword Set while preserving scientific rigor.
This highlights \textit{Spacer}'s potential to provide scientific inspirations not limited to simple extensions of the current art.
Further examples are available in our \href{https://github.com/Asteromorph-Corp/Spacer}{Github repository}.


\subsection{Restoring Calcium Oscillations in Hepatocellular Carcinoma}

\def\promptfontsize{\scriptsize}
\begin{center}
    \begin{tcolorbox}[
            breakable,
            title={ },
            colback=blackbg,
            colframe=blackheader
        ]
        \begin{lstlisting}[
breaklines=true,
basicstyle=\ttfamily\promptfontsize,
columns=fullflexible,
breakindent=0pt,
frame=none,
breakatwhitespace=true,
mathescape=true
]
{
  "concept": "Controlled noise injection into calcium signaling pathways may restore oscillatory coherence in hepatocellular carcinoma cells through stochastic resonance mechanisms. This approach could potentially re-establish calcium-dependent cell cycle checkpoint regulation that appears disrupted in malignant hepatocytes compared to normal physiological fluctuation patterns.",
  "supporting_dois": [
    "10.1002/ijc.21591",
    "10.1002/jcb.1140",
    "10.1002/jnr.1120",
    180 more DOIs...
  ],
  "rationale": [
    "Hepatocellular carcinoma cells exhibit fundamentally altered calcium homeostasis and signaling dynamics compared to healthy hepatocytes. These cancer cells show disrupted calcium oscillations, altered NF$\kappa$B signaling patterns, decreased expression of plasma membrane Ca2+-ATPase leading to sustained high intracellular calcium levels, and disrupted circadian rhythms that normally coordinate calcium oscillations with cell cycle checkpoints.",
    "Stochastic resonance represents a fundamental principle where introducing optimal levels of noise enhances temporal coherence of weak periodic signals in nonlinear systems. In biological contexts, controlled introduction of specific noise patterns can amplify oscillatory coherence, and non-Gaussian noise patterns have been demonstrated to optimize temporal patterns in calcium signaling networks.",
    "Hepatocyte calcium signaling networks possess the essential nonlinear dynamics required for stochastic resonance effects. Theoretical models demonstrate that intracellular calcium systems in sub-threshold, excitable states can exhibit coherence resonance, where optimal noise levels enhance regularity of intrinsic oscillations. These systems approximate real cell dimensions necessary for meaningful stochastic resonance effects.",
    "Calcium oscillation coherence serves as a critical signaling code that encodes information determining activation of specific downstream genetic programs controlling cell cycle progression. The precise temporal patterns of these oscillations are decoded by pathways that differentially activate transcription factors like NF-$\kappa$B and NFAT, which regulate expression of key cell cycle checkpoint proteins such as p21 and cyclin D.",
    "Cancer cells demonstrate selective vulnerability to calcium-based interventions due to their altered calcium handling mechanisms. Targeted approaches can exploit differential calcium transport, sensing receptor expression, and homeostasis between healthy and malignant cells to selectively inhibit proliferation or induce calcium-mediated cell death pathways like calcioptosis.",
    "Controlled aperiodic modulation of extracellular calcium concentrations provides a technologically feasible method to inject noise into intracellular calcium signaling systems. Modulating extracellular calcium directly influences intracellular oscillation dynamics by altering electrochemical gradients and flux through ion channels and pumps, enabling systematic enhancement of signal-to-noise ratios in endogenous oscillations.",
    "The proposed experimental parameters for calcium modulation fall within safe, physiologically relevant ranges for healthy hepatocytes while effectively targeting cancer cell vulnerabilities. Extracellular concentrations of 0.8-1.25 mM, modulation frequencies of 0.05-1.0 Hz, and pulse durations of 1-5 seconds align with natural cellular signaling kinetics while avoiding cytotoxicity in normal cells.",
    "Advanced microfluidic technologies combined with sophisticated live-cell imaging enable precise delivery of complex temporal noise patterns and quantitative measurement of their effects. These systems provide the necessary temporal resolution and spatial control to generate non-Gaussian noise patterns while monitoring calcium oscillation coherence and cell cycle status in real-time.",
    "Restoring physiological-like coherence to disrupted calcium oscillations in cancer cells may re-engage homeostatic control mechanisms and suppress malignant phenotypes. The transition from healthy to cancerous states often involves loss of coherent signaling dynamics in favor of simplified, persistent signaling states that promote proliferation. Therefore, re-imposing coherent, physiological-like signal patterns could restore homeostatic regulatory network function."
  ]
}
\end{lstlisting}
    \end{tcolorbox}
    \captionof{figure}{\textit{Spacer}'s output of \textit{Restoring Calcium Oscillations in Hepatocellular Carcinoma}. Some supporting DOIs are omitted for brevity.}\label{fig:result-salad76}
\end{center}

\begin{figure}[htb]
    \centering
    \includegraphics[width=0.80\textwidth]{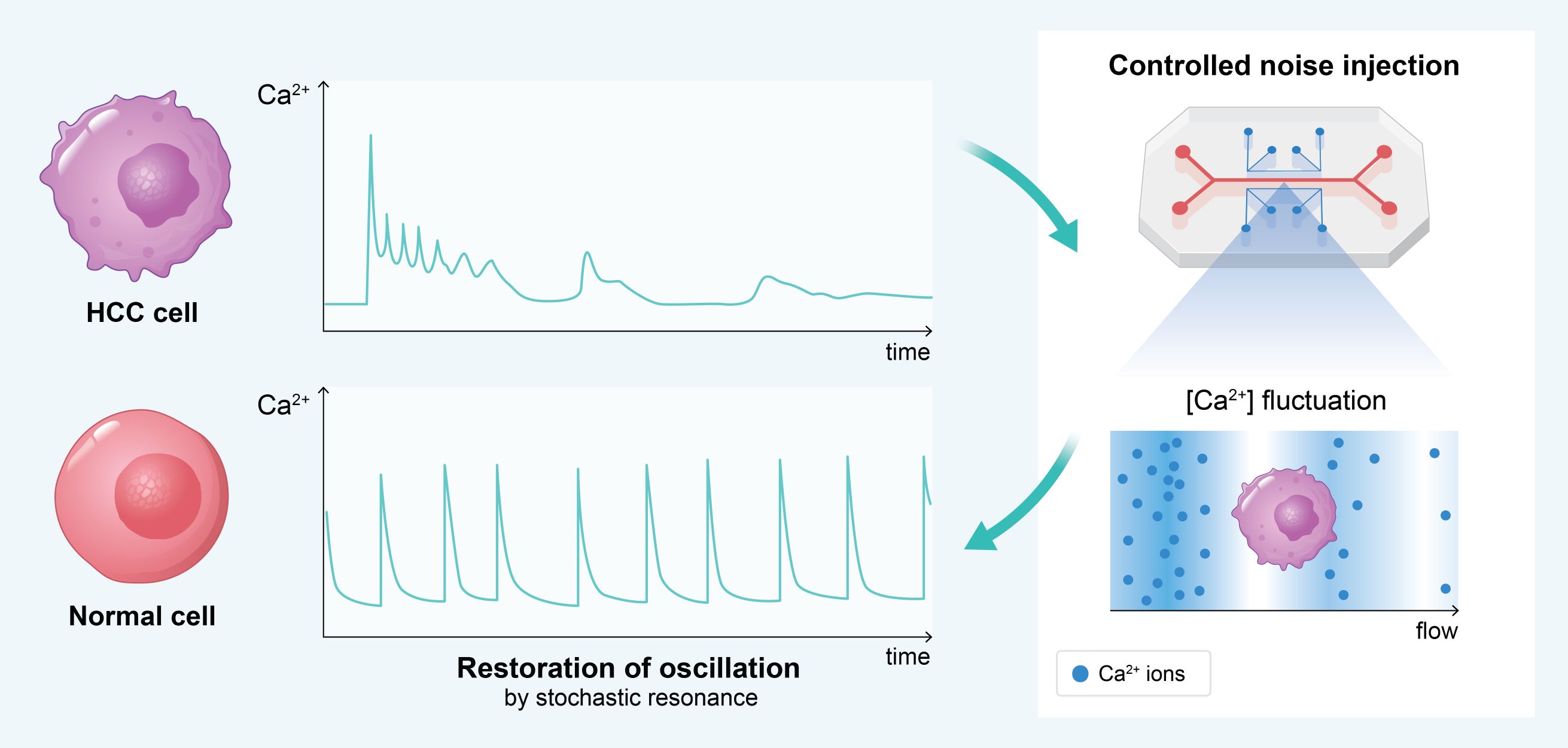}
    \caption{Hepatocellular carcinoma cells exhibit disrupted calcium oscillations. Controlled noise injection as extracellular calcium fluctuation could restore oscillatory coherence toward normal state, suppressing malignant phenotype.}
    \label{fig:scheme-salad76}
\end{figure}

Calcium ions are essential mediators in various cellular signaling pathways.
In particular, calcium oscillations are integral to cellular processes including the regulation of cell cycle progression and NFAT activation \cite{sneyd2017dynamical, kapur2007cell, kar2016control}.
In malignant hepatocytes, calcium signaling undergoes extensive dysregulation in which calcium oscillations become impaired and cell cycle checkpoints are compromised \cite{sun2019central, tang2015cd147}.

\textit{Spacer} proposes to suppress malignant phenotypes of hepatocellular carcinoma cells by restoring calcium oscillatory coherence with stochastic resonance.
Stochastic resonance is a phenomenon in nonlinear systems where the addition of an appropriate level of noise can enhance oscillatory coherence \cite{zhang2008selective}.
When applied to calcium signaling, controlled noise injection could potentially restore the temporal coherence of disrupted oscillations \cite{li2005internal, gong2010non, lang2008roles}.
Such noise can be introduced through methods like extracellular calcium concentration modulation or microfluidic technologies \cite{ding2011forster, chang2014parallel, komen2020controlled}.
Restoring calcium oscillation coherence could in turn recover normal calcium-dependent checkpoint regulation, thereby suppressing malignant phenotypes.
Furthermore, this approach could be extended to selective targeting based on the differential sensitivity of cancer cells to calcium perturbation \cite{wang2019cancer, bai2024bioinspired}.

This example suggests a therapeutic approach of applying stochastic resonance, which has not been well-established in cancer treatment.
The integration of calcium dynamics, nonlinear systems theory, and cellular oncology demonstrates the capability of connecting disparate domains.


\subsection{ATP Allocation Patterns Predict Cellular State Transitions}

\def\promptfontsize{\scriptsize}
\begin{center}
    \begin{tcolorbox}[
            breakable,
            title={ },
            colback=blackbg,
            colframe=blackheader
        ]
        \begin{lstlisting}[
      breaklines=true,
      basicstyle=\ttfamily\promptfontsize,
      columns=fullflexible,
      breakindent=0pt,
      frame=none,
      breakatwhitespace=true,
      mathescape=true
    ]
{
  "concept": "Cellular state transitions can be predicted by quantifying ATP allocation patterns across competing metabolic pathways. This approach treats cells as hierarchical energy-processing systems where information emerges from metabolite flux distributions during differentiation or stress responses. ATP-dependent processes actively maintain cellular organization and non-equilibrium states, directly linking energy budgets to fundamental cellular properties.",
  "supporting_dois": [
    "10.1002/bies.200900057",
    "10.1002/btpr.448",
    "10.1002/jbmr.3390",
    126 more DOIs...
  ],
  "rationale": [
    "Cellular state transitions are fundamentally governed by ATP redistribution across competing metabolic pathways. During differentiation or stress responses, cells systematically reallocate their energy budgets, creating predictable patterns that reflect underlying bioenergetic constraints and regulatory priorities.",
    "Cells function as hierarchical energy-processing systems where metabolic networks exhibit natural flux ordering. Upstream reaction rates constrain downstream fluxes, creating quantifiable hierarchies of reaction importance that generate predictive information through reduced metabolic entropy and selective pathway activation.",
    "ATP allocation is dynamically controlled through rapid allosteric regulation and feedback mechanisms. Enzymes like pyruvate kinase M2 and regulatory systems like AMPK enable near-instantaneous metabolic shifts within seconds to minutes, while transcriptional control governs slower transitions during differentiation processes.",
    "Predictive models can be constructed by integrating thermodynamic constraints with kinetic parameters and stochastic methods. This approach incorporates Gibbs free energy changes, cofactor ratios, and trade-off analysis to ensure biophysically plausible flux distributions that account for biological variability.",
    "Quantitative experimental validation is achievable through multi-modal approaches combining fluxomics, metabolomics, and live-cell imaging. Carbon-13 flux analysis, real-time ATP indicators, and mass spectrometry provide the temporal resolution and quantitative precision needed to parameterize and validate predictive models.",
    "This framework advances metabolic engineering by identifying energetic bottlenecks and targetable regulatory nodes. The approach surpasses traditional flux balance analysis by incorporating rapid regulatory mechanisms and provides evolutionarily validated insights into conserved bioenergetic strategies across lineages."
  ]
}
    \end{lstlisting}
    \end{tcolorbox}
    \captionof{figure}{\textit{Spacer}'s output of \textit{ATP Allocation Patterns Predict Cellular State Transitions}. Some supporting DOIs are omitted for brevity.}\label{fig:result-salad60}
\end{center}

\begin{figure}[htb]
    \centering
    \includegraphics[width=0.90\textwidth]{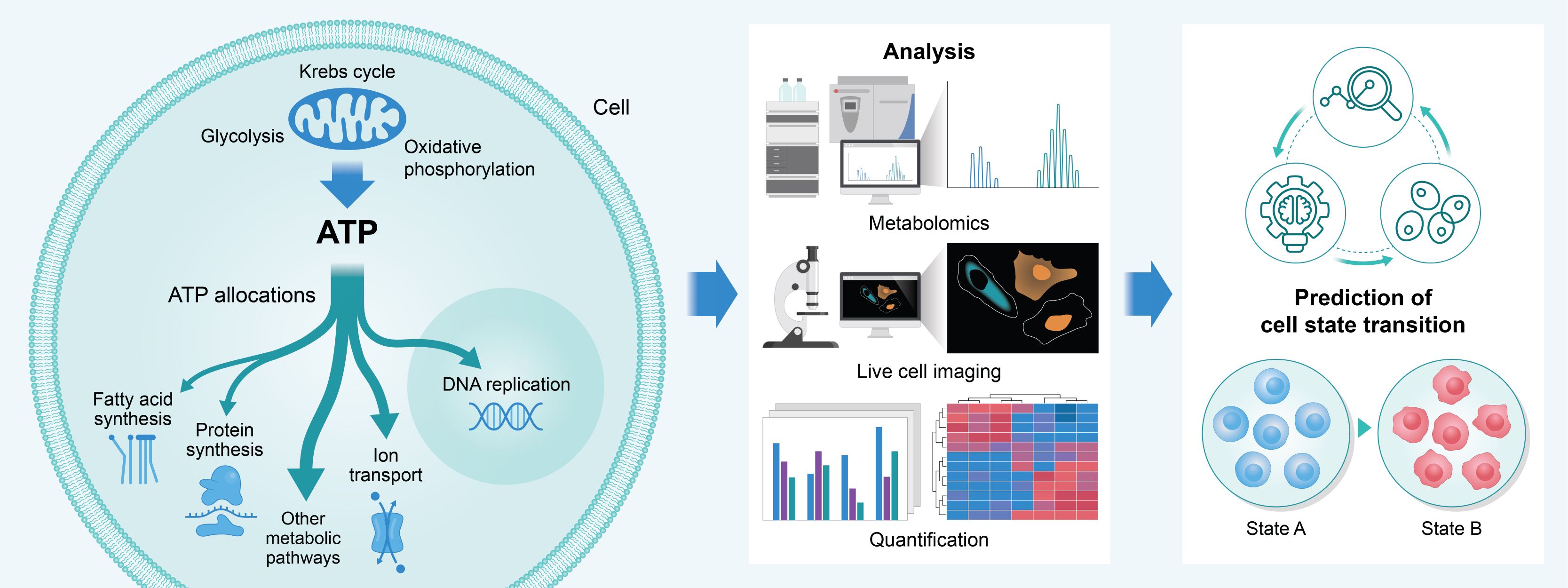}
    \caption{ATP is distributed across diverse metabolic pathways. Quantifying the allocation may enable prediction of cellular state transitions.}
    \label{fig:scheme-salad60}
\end{figure}

\begingroup
\sloppy

Adenosine triphosphate (ATP) is an universal energy currency in cells.
The flow of ATP through metabolic pathways is dynamically regulated based on functional priorities and energy demands.
Thus, ATP distribution patterns can provide predictive information about a cell's current state and potential transitions \cite{yang2021physical}.
\textit{Spacer} suggests a framework for the prediction of cellular state transitions based on quantifying ATP allocations.
Changes in energy deployment can be tracked through multiple experimental approaches, including fluxomics, metabolomics, and live-cell imaging techniques \cite{mookerjee2017quantifying, lobas2019genetically, xiao2025genetically, schaub2008metabolic}.
These methods provide the temporal resolution and quantitative precision needed to monitor dynamic metabolic shifts in real-time.

This framework presents a new perspective in cell state modeling that contributes to both conventional cell biology and cutting-edge research areas.
Recent advances in single-cell foundation models and their extension toward virtual cells have largely centered on single cell expression modalities such as scRNA-seq and CITE-seq, which predominantly provide static snapshots \cite{cui2024scgpt, heimberg2025cell, zeng2025cellfm, adduri2025predicting}.
By incorporating ATP allocation patterns, these approaches can encode mechanistic energy budgets alongside molecular features.
Such an intergration of energy landscape with molecular data may extend the application from static depiction to dynamic cell state trajectory prediction.

\endgroup

\clearpage


\subsection{Overexpressing Olfactory Receptors for Gut Microbiome Control}

\def\promptfontsize{\scriptsize}
\begin{center}
    \begin{tcolorbox}[
            breakable,
            title={ },
            colback=blackbg,
            colframe=blackheader
        ]
        \begin{lstlisting}[
      breaklines=true,
      basicstyle=\ttfamily\promptfontsize,
      columns=fullflexible,
      breakindent=0pt,
      frame=none,
      breakatwhitespace=true,
      mathescape=true
    ]
{
  "concept": "Intestinal epithelial cells may be engineered to overexpress olfactory receptors that detect specific microbial metabolites, potentially coupling these repurposed chemosensory signals to NF$\kappa$B pathways for localized antimicrobial peptide secretion. This approach could enable spatially-restricted epithelial signaling nodes that autonomously modulate regional microbial populations based on real-time metabolite detection.",
  "supporting_dois": [
    "10.1002/cbdv.200890099",
    "10.1002/fft2.301",
    "10.1016/j.isci.2018.05.009",
    146 more DOIs...
  ],
  "rationale": [
    "Olfactory receptors demonstrate proven capacity for microbial metabolite detection in non-nasal tissues. These G-protein coupled receptors are naturally expressed in intestinal epithelium and respond to bacterial compounds including short-chain fatty acids through specific receptors like OR2T4, OR10S1, and Olfr78. Their rapid signaling kinetics enable real-time metabolite sensing within biologically relevant timeframes.",
    "AAV-mediated gene delivery systems provide robust technical feasibility for stable receptor overexpression in intestinal epithelium. AAV vectors achieve efficient transduction with minimal immunogenicity, while inducible expression systems and cell-specific promoters ensure controlled, targeted transgene expression that addresses epithelial turnover challenges.",
    "GPCR-to-NF$\kappa$B signaling pathways establish mechanistic linkage between metabolite detection and antimicrobial response. G-protein subunits activated by olfactory receptors can initiate cascades through phospholipase C and protein kinase C that converge on NF$\kappa$B activation. The oscillatory dynamics of NF$\kappa$B signaling enable precise temporal control over antimicrobial peptide production.",
    "Short-chain fatty acids and indole derivatives represent optimal target metabolites for engineered detection systems. SCFAs reach mmol/L concentrations during bacterial overgrowth with well-characterized spatial gradients, while indole metabolites activate aryl hydrocarbon receptors to modulate immunity with minimal off-target effects. Both metabolite classes demonstrate favorable safety profiles and physiological relevance.",
    "Spatially-restricted signaling architecture enables precise microbial modulation while preserving gut homeostasis. This approach targets dysbiotic regions without disrupting commensal diversity, based on compartmentalized immunity principles. Engineered responses targeting specific metabolites may reduce antimicrobial resistance emergence compared to broad-spectrum interventions.",
    "Advanced organoid and microfluidic validation platforms provide robust empirical testing capabilities. Human intestinal organoids coupled with gut-on-a-chip technologies enable precise monitoring of metabolite-induced receptor activation and spatially-resolved analysis of antimicrobial peptide secretion, supporting translational development."
  ]
}
  \end{lstlisting}
    \end{tcolorbox}
    \captionof{figure}{\textit{Spacer}'s output of \textit{Overexpressing Olfactory Receptors for Gut Microbiome Control}. Some supporting DOIs are omitted for brevity.}\label{fig:result-salad55}
\end{center}

\begin{figure}[htb]
    \centering
    \includegraphics[width=0.70\textwidth]{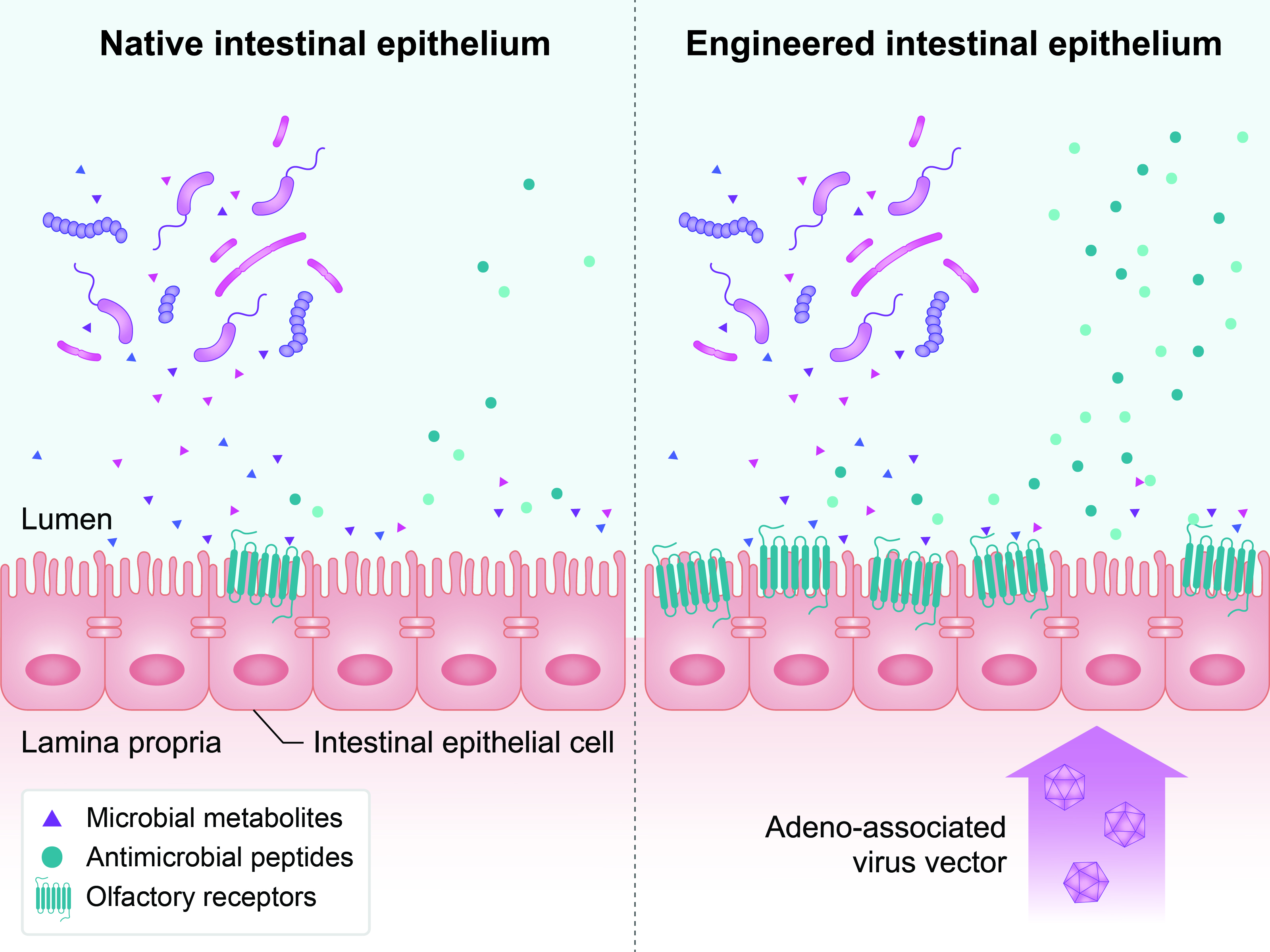}
    \caption{Intestinal epithelial cells can be engineered to overexpress olfactory receptors. This may lead to localized antimicrobial peptide secretion, enhancing intestinal microbial regulation.}
    \label{fig:scheme-salad55}
\end{figure}

\begingroup
\sloppy

Olfactory receptors are widely recognized for their role in odor detection within the nasal epithelium.
They are also naturally expressed in the intestinal epithelium, where they function as chemosensors for bacterial metabolites \cite{pluznick2013olfactory, lednovich2019or31, yasi2019rapid}.
These receptors may influence NF-κB signaling pathways, leading to cascades that trigger NF-κB mediated antimicrobial pepdite (AMP) secretion \cite{wu2021activation, peng2024nad, stockinger2014trif}.
This creates an autonomous, metabolite-responsive defense system that operates in real-time based on local microbial activity.
\textit{Spacer} argues for engineering intestinal epithelial cells to overexpress olfactory receptors, triggering AMP secretion in response to their stimulation.
Combining adeno-associated virus (AAV) vectors with cell-specific promoters enables the selective delivery of transgenes to the intestinal epithelium \cite{polyak2012identification, chan2021engineering, yuan2022mitigating}.

The proposed concept introduces a therapeutic approach that connects intestinal olfactory receptors to the regulation of AMP secretion.
Built upon well characterized individual components, such as intestinal olfactory receptors and AMP secretion, their integration into a spatially confined, metabolite-responsive immune circuit has not yet been explored.
This strategy could enable spatially restricted immune modulation to preserve beneficial commensal communities and reduce the risk of antibiotic resistance emergence.

\endgroup

\clearpage
\section{Validations\label{sec:experiments}}

In validation, we examined three core parts of our work: (a) \textsc{Nuri}'s effectiveness at generating Keyword Sets resembling those of high-impact papers (\cref{subsec:nuri_validation}), (b) \textsc{Weaver}'s capability to find concepts in Keyword Sets (\cref{subsec:paper_reconstruction}), and (c) \textit{Spacer}'s end-to-end performance in comparison to experts and state-of-the-art LLMs (\cref{subsec:embedding}).

As \textsc{Nuri}'s core approach is to search for sets with an estimated high impact, the validation of (a) focuses primarily on the accuracy of its estimator.
The remaining two validations merits further discussion.

Given that both (b) and (c) require a quality assessment of Statements, one might be tempted to directly score each sample's quality with LLM or expert human judges, as done in \cite{ruan2025liveideabenchevaluatingllmsdivergent,gao2025graphaiideasleveraging,si2024llmsgeneratenovelresearch}.
However, expert judge scores suffer from high variance arising from personal differences.
Judgements from LLMs, on the other hand, are easily confounded by contextual and stylistic clues (\cref{subsubsec:scaffolding}), and suffer from a bias to prefer LLM outputs \cite{laurito2024ai}.
Certainly, these issues can and should be mitigated by normalizing the data, but we argue that the approach is fundamentally misguided.
The value of research ideas ultimately comes from the results of their execution; being able to reliably estimate their result would mean the idea is already obvious and thus useless.
It is unsurprising, then, that any \textit{a priori} evaluation of research ideas will suffer from inherent variance and bias.

We propose to sidestep this problem entirely by using existing high-quality published human research.
Since the theses of these papers are then verified to yield high-impact results, we can use semantic similarity to measure the performance of systems aiming to create research ideas.

Even with this approach, data must be compared in the same form and style, with content irrelevant to the evaluation removed.
Since we aim to only evaluate ideas, we redacted the experimental results from abstracts of existing research papers, while converting \textit{Spacer}'s Statements from their structured JSON form into prose.
We call the resulting paragraph containing the core hypothesis a \textit{thesis paragraph}.


\newcommand{\f}{f_{(\cdot)}}
\newcommand{\nf}{\operatorname{EVAL}}

\subsection{Nuri\label{subsec:nuri_validation}}

We continue using the notation of the description of \textsc{Nuri} in \Cref{subsubsec:nuri}.
To evaluate the explanatory power of $\f$, we prepare a validation set $\mathbb{P}$ with 180,000 papers in peer-reviewed biology journals.
Since our assessment must be causal, when evaluating a paper $p$, we must not take into account papers published after it.
We therefore write $\mathbb{P}_{<p}$ for the subset of all papers published before the paper $p$.
Then $\nf(p)\coloneq f_{\mathbb{P}_{<p}}(\kw(p))$ is \textsc{Nuri}'s estimate of the impact a paper with the Keyword Set $\kw(p)$ would have had, given the knowledge base prior to it.
By comparing this value to the realized significance of the paper $\fwci(p)$, we can judge the accuracy of $\f$.

\begin{figure}[htb]
    \centering
    \includegraphics[width=0.6\textwidth]{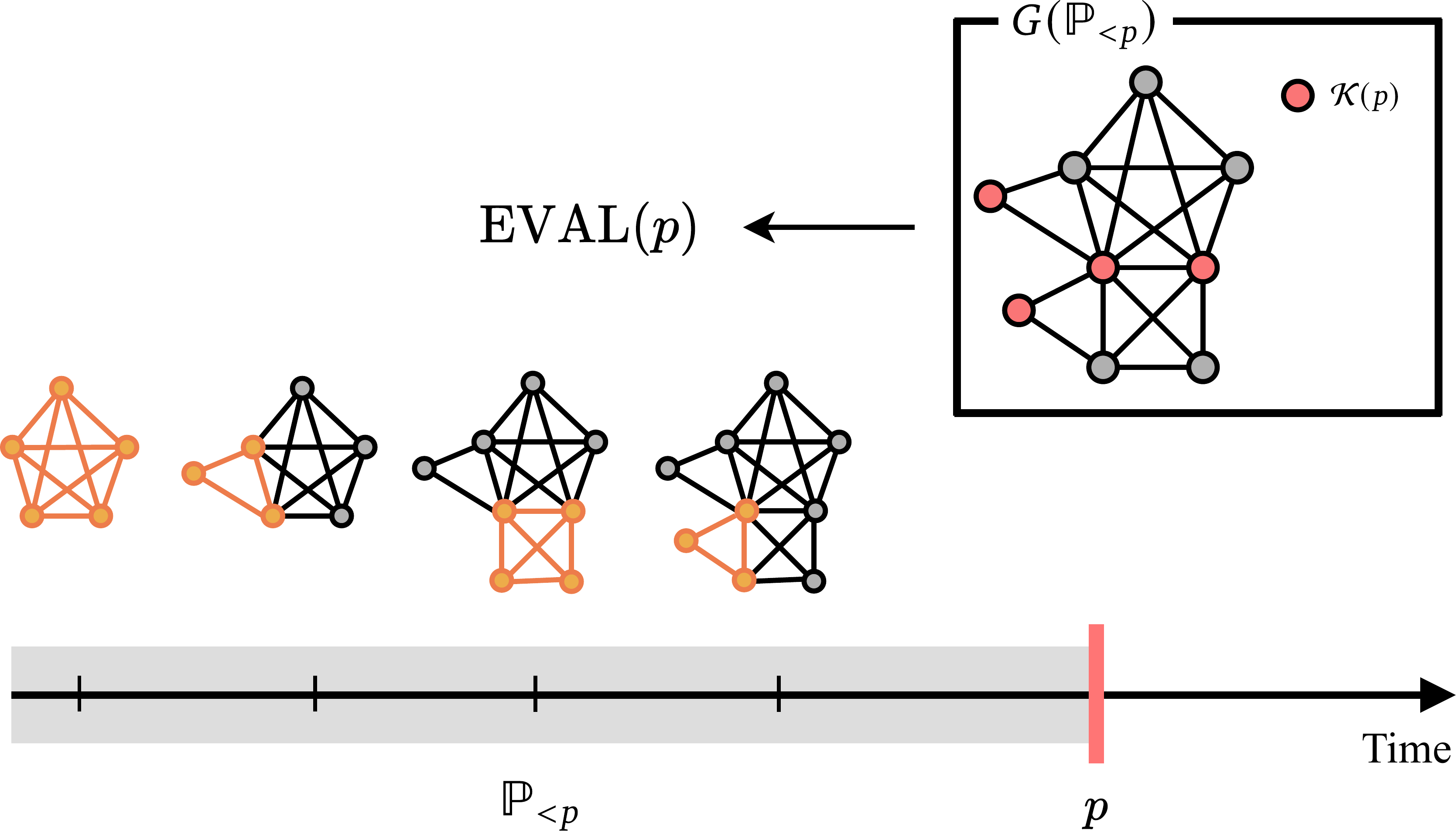}
    \caption{Schematic of the evaluation process of $\f$. For each paper, a complete graph is constructed with its keywords as vertices. $G(\mathbb{P}_{<p})$ is formed by merging graphs corresponding to all papers published before $p$.}
    \label{fig:nuri_evaluation}
\end{figure}

The performance of $\f$ was measured by using it on a downstream binary classification task.
We chose 400 papers to be used in the task: 200 high-impact papers with $\fwci(p) \geq 15$ and 200 low-impact papers with $\fwci(p) < 1$.
For comparison of the distributions, we plotted a Receiver Operating Characteristic (ROC) curve of $\nf(p)$ and $\fwci(p)$ by calculating the sensitivity and 1-specificity for varying thresholds.

\begin{figure}[htb]
    \centering
    \includegraphics[width=0.85\textwidth]{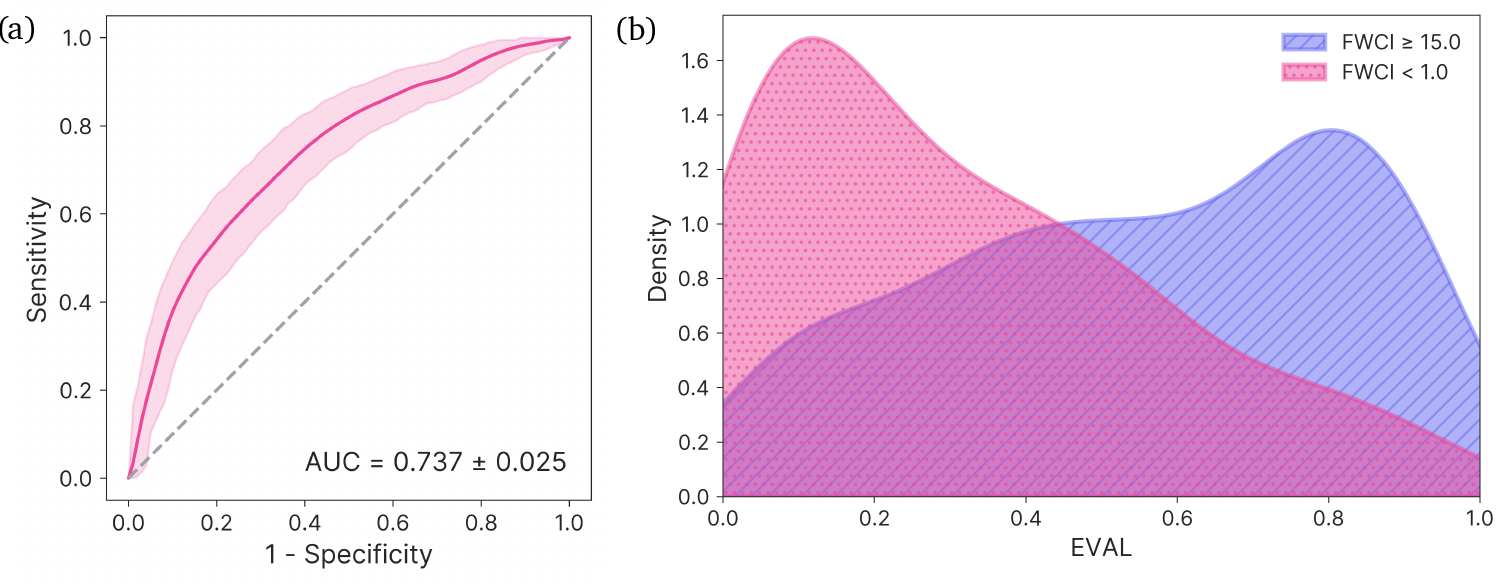}
    \caption{
        Performance of $\nf$ on a validation set of 200 high-impact and 200 low-impact papers.
        (a) ROC curve of the function $\nf$ for predicting high-impact papers, with an area under the curve (AUC) value of $0.737 \pm 0.025$.
        The red solid line represents the classification performance of $\nf$, with its 95\% confidence interval shaded around.
        The gray dashed line represents the classification performance of a random classifier (AUC = 0.5).
        (b) Distribution of $\nf$ scores for the high (blue) and low (red) impact papers, showing clear separation between the two classes.
        Both curves are normalized to have unit area.
    }
    \label{fig:nuri_result}
\end{figure}

\begin{figure}[htb]
    \centering
    \includegraphics[width=0.95\textwidth]{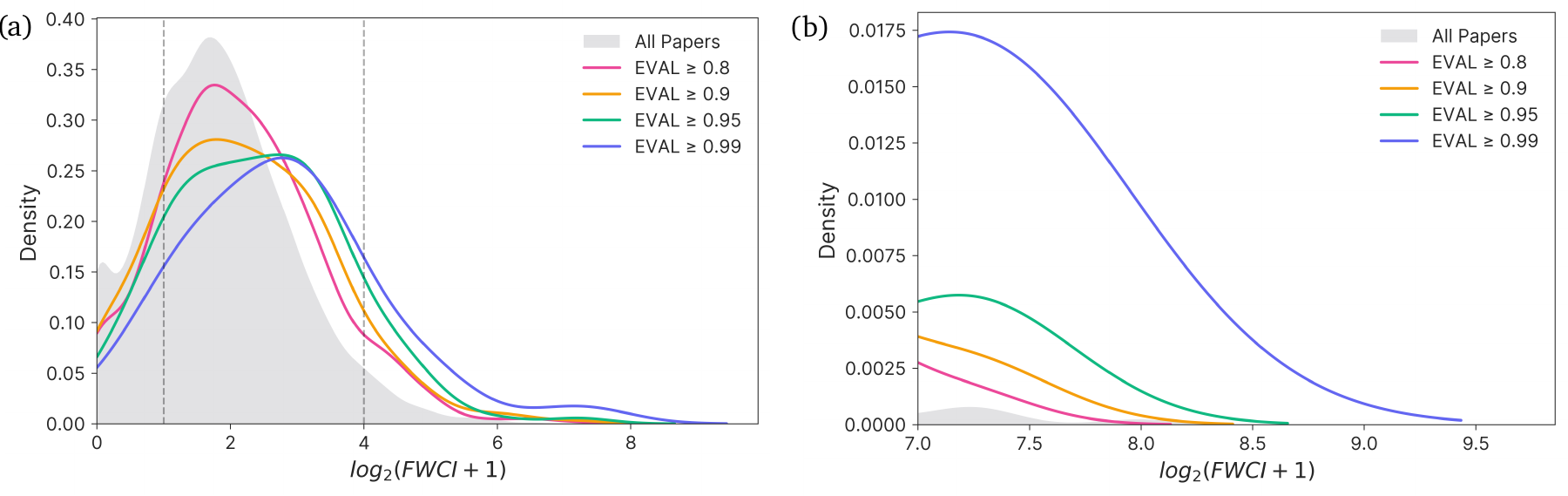}
    \caption{
        Distribution of $\log_2(\fwci + 1)$ for a set of 10,000 randomly sampled papers.
        The gray shaded area represents the distribution for all papers in the set.
        The colored lines represent the distribution for the subset of papers whose $\nf$ exceeds a specific threshold: $\nf \geq 0.8$ (pink), $\nf \geq 0.9$ (orange), $\nf \geq 0.95$ (green), and $\nf \geq 0.99$ (blue).
        Each curve is normalized to have unit area.
        The vertical dashed lines indicate the cutoff values for high and low-impact papers.
        (a) Full-range plot ($0 \leq \log_2(\fwci+1) \leq 10$).
        (b) Truncated plot ($7 \leq \log_2(\fwci+1) \leq 10$).
    }
    \label{fig:nuri_test_result}
\end{figure}

The results of the classification task are plotted in \cref{fig:nuri_result}, which demonstrates that $\f$ effectively distinguishes between high and low impact papers.
We also examined the distribution of $\log_2(\fwci + 1)$ values for a set of 10,000 randomly sampled papers across different $\nf(p)$ thresholds (Figure~\ref{fig:nuri_test_result}).
Notably, higher $\nf(p)$ thresholds lead to a substantial increase in the proportion of papers with very high citation impact.

\begin{figure}[htb]
    \centering
    \includegraphics[width=0.32\textwidth]{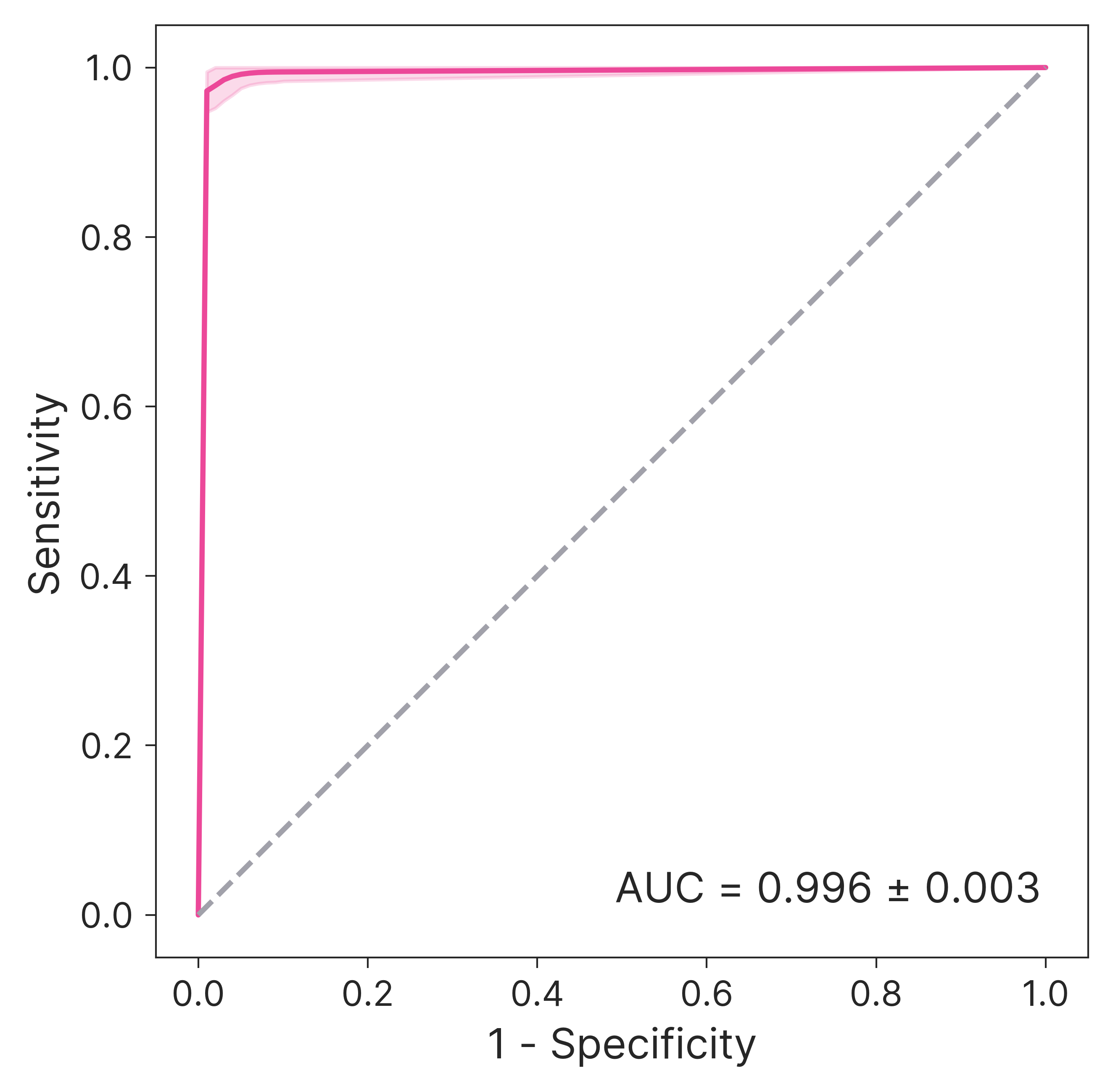}
    \caption{
        ROC curve of the function $\nf$ for predicting whether the Keyword Set is extracted from a paper or is randomly selected.
        The red solid line represents the classification performance of $\nf$, with its 95\% confidence interval shaded around.
    }
    \label{fig:roc_curve_rand}
\end{figure}

As a further sanity check, we applied $f_{\mathbb{P}}$ to differentiate paper-originated Keyword Sets from randomly selected sets.
The two groups displayed a marked divergence in their $f_{\mathbb{P}}$ values, where the AUC of this classification task reached $0.996 \pm 0.003$, as shown in Figure~\ref{fig:roc_curve_rand}.

As \textsc{Nuri} was designed to search for sets with high $f_{\mathbb{P}}$ values, the accuracy of $\f$ directly translates into potential of the resulting Keyword Sets.
Thus, our experiments serve as strong evidence of \textsc{Nuri} being able to create Keyword Sets carrying the necessary initiative to direct the next stages of our pipeline to meaningful research.


\subsection{Reconstructions of Latest Cutting-edge Scientific Concepts\label{subsec:paper_reconstruction}}

As discussed in \cref{sec:experiments}, instead of scoring \textsc{Weaver}'s outputs on \textsc{Nuri}-generated Keyword Sets, we test whether it can reconstruct the research thesis of existing papers given an extracted Keyword Set.
To preclude the possibility that the base model already has knowledge of the papers, we randomly chose 158 abstracts of papers published in acclaimed scientific journals including Science and Nature after May 1st, 2025.
We then converted these abstracts into thesis paragraphs, and extracted Keyword Sets.
These sets were then processed through \textsc{Weaver}, in place of the \textsc{Nuri}-generated Keyword Sets, to create the reconstructed thesis paragraph for comparison.

Table~\ref{tab:paper_reconstruction_example} presents side-by-side comparisons of original theses and those reconstructed by \textsc{Weaver}.
Across all examples, \textsc{Weaver} precisely recovers the original ideas, in particular preserving the logical structure and domain-specific subtleties.

For a quantitative comparison, we implemented a test comparing the original ideas and \textsc{Weaver} reconstructions using o3.
We judged the similarity of the two inputs in five aspects: logic, topic, objective, approach, and an ``overall'' decision.
Each of the five prompts used are available in Appendix~\ref{app:judgeprompt}.
We summarize the results in Table~\ref{tab:reconresults}, which show that reconstructed paragraphs consistently pass at a rate of over 85\% across all journals used.
To view all 158 papers and their original \& reconstructed thesis paragraphs, refer to our \href{https://github.com/Asteromorph-Corp/Spacer}{Github repository}.

\begin{table}[htb]
    \centering
    \caption{Pass rates of reconstructed theses based on similarity to the original theses. Journals classified as Others include Nature Methods, Cell, and Neuron. The count of passed theses over paper count is displayed by category, with the total accumulated percentage on the rightmost column.}
    \begin{tabular}{ccccc|c}
        \toprule
        \multirow{2}{*}{Criteria} & \multicolumn{5}{c}{Pass Rates}                                                         \\
        \cmidrule{2-6}
                                  & Science                        & Nature & Others & \multicolumn{2}{c}{Total}           \\
        \midrule
        Logic                     & 75/78                          & 47/50  & 30/30  & 152/158                   & 96.20\% \\
        Topic                     & 78/78                          & 50/50  & 30/30  & 158/158                   & 100.0\% \\
        Objective                 & 77/78                          & 48/50  & 30/30  & 155/158                   & 98.10\% \\
        Approach                  & 76/78                          & 48/50  & 30/30  & 154/158                   & 97.47\% \\
        Overall                   & 67/78                          & 40/50  & 28/30  & 135/158                   & 85.44\% \\
        \bottomrule
    \end{tabular}
    \label{tab:reconresults}
\end{table}

Overall, we have exhibited that \textsc{Weaver} can reconstruct research theses crafted and peer-reviewed by human experts from a sparse set of keywords.
We take this as strong evidence that innovative research can be inspired from these Keyword Sets, and that \textsc{Weaver} already shows potential to do so.

\begin{landscape}
    \begin{table}[ht]
        \centering
        \caption{Results of paper reconstruction using Weaver. From left to right: original theses from the papers, extracted Keyword Sets, theses reconstructed by Weaver, and source references.}
        {
            \scriptsize
            \begin{tabularx}{\linewidth}{Xp{0.15\linewidth}Xc}

                \toprule
                \textbf{Original thesis}                                                                                                                                                                                                                                                                                                                                                                                                                                                                                                                                                                                                                                                                                                                                                                                                                                                                                                                                                                                                                                                                                                                            & \textbf{Keyword Set}                                                                                                                                                                                                                                                                    & Weaver\textbf{-generated thesis}                                                                                                                                                                                                                                                                                                                                                                                                                                                                                                                                                                                                                                                                                                                                                                                                                                                                                                                  & \textbf{Source}       \\
                \midrule
                Chimeric antigen receptor T cell therapy in solid tumors faces limitations including immunosuppression and antigen heterogeneity. While armoured CAR T cells engineered to secrete proinflammatory cytokines like IL-12 and IL-2 have been developed to enhance efficacy, their clinical application has been restricted by toxicity arising from transgene expression in peripheral tissues. To address this, an innovative CRISPR knock-in strategy ensures transgene expression is localized specifically to the tumor microenvironment by leveraging specific endogenous genes, such as NR4A2 and RGS16, which exhibit tumor-restricted expression. This approach, demonstrated to improve CAR T cell polyfunctionality and activate endogenous antitumour immunity in syngeneic and xenogeneic models, enhances the therapeutic index by driving cytokine delivery directly to the tumor site, thereby improving antitumour efficacy and survival while maintaining a favorable safety profile. This method is applicable to patient-derived CAR T cells.                                                                                      & CAR T cells, solid tumors, immunosuppression, antigen heterogeneity, IL-12, IL-2, toxicity, CRISPR knock-in, tumor microenvironment, NR4A2, RGS16                                                                                                                                       & Chimeric antigen receptor (CAR) T cells, which target specific tumor antigens, face challenges in treating solid tumors due to immunosuppression and tumor antigen heterogeneity. To address these barriers, a novel approach proposes engineering CAR T cells to locally deliver interleukin (IL)-12 and IL-2 within the tumor microenvironment. Additionally, CRISPR knock-in technology is employed to knock out the NR4A2 and RGS16 genes in these CAR T cells. This combined strategy aims to overcome immunosuppression and tumor antigen heterogeneity, enhancing the efficacy of CAR T cell therapy in solid tumors.                                                                                                                                                                                                                                                                                                                      & \cite{chen2025-hi}    \\
                \midrule
                The enteric nervous system must detect luminal nutrients to control digestive processes, yet the specific neuronal pathways mediating this sensory function remain undefined along with the system's capacity to distinguish between different chemicals. Calcium imaging can reveal which enteric pathways activate when different nutrients contact the mouse jejunum, potentially showing that distinct neurochemically defined ensembles of myenteric and submucosal neurons respond to specific luminal substances. Since enteric neurons likely detect nutrients indirectly through epithelial intermediates rather than direct nutrient sensitivity, the epithelium probably transduces luminal chemical information via specific signaling molecules like serotonin to communicate with different neuronal populations. This nutrient detection system may organize spatially along the intestine's radial axis, with signals potentially flowing from villus epithelium to myenteric plexus and then to submucosal plexus in a defined sequence that reflects the anatomical relationship between luminal proximity and neural processing. & enteric nervous system, luminal nutrients, neuronal pathways, calcium imaging, mouse jejunum, neurochemically defined ensembles, myenteric neurons, submucosal neurons, serotonin, epithelium, signaling molecules, radial axis, villus epithelium, myenteric plexus, submucosal plexus & The enteric nervous system (ENS) processes luminal nutrients through distinct neuronal pathways. This research idea proposes that specific neurochemically defined ensembles of enteric neurons are activated by luminal nutrients. The experimental approach employs calcium imaging to visualize these activation patterns in live mouse jejunum. Myenteric neurons and submucosal neurons, two major classes of enteric neurons, are considered. The investigation focuses on how serotonin, released from the epithelium upon luminal nutrient exposure, acts as a signaling molecule. This approach examines the activation of myenteric neurons and submucosal neurons along the radial axis of the intestinal wall. The study specifically addresses how these neuronal populations are activated by signaling molecules released from the villus epithelium, which lies in close proximity to the myenteric plexus and submucosal plexus. & \cite{fung2025-rl}    \\
                \midrule
                Adult hippocampal neurogenesis is known to be involved in memory formation and mood regulation. These processes are indicative of neurogenesis in humans, with proliferating neural progenitor cells identifiable within the dentate gyrus. Given this, the persistence and characteristics of adult hippocampal neurogenesis in humans, beyond early childhood, can be elucidated by systematically analyzing neural progenitor cell stages across different human age groups using single-nucleus RNA sequencing.                                                                                                                                                                                                                                                                                                                                                                                                                                                                                                                                                                                                                                 & hippocampal neurogenesis, memory formation, mood regulation, neural progenitor cells, dentate gyrus, adults, age groups, single-nucleus RNA sequencing, human                                                                                                                           & Hippocampal neurogenesis, the process of generating new neurons in the hippocampus, plays a crucial role in memory formation and mood regulation. This process involves neural progenitor cells within the dentate gyrus, a specific hippocampal subregion. While hippocampal neurogenesis persists throughout life in humans, its rate and functional significance in adults across different age groups remain unclear. A study aims to address this by utilizing single-nucleus RNA sequencing to profile neural progenitor cells in the dentate gyrus of human adults spanning various age groups.                                                                                                                                                                                                                                                                                                                                            & \cite{dumitru2025-fp} \\
                \bottomrule
            \end{tabularx}
        }
        \label{tab:paper_reconstruction_example}
    \end{table}
\end{landscape}


\subsection{Sentence Embeddings Analysis\label{subsec:embedding}}

We next evaluate the end-to-end performance of \textit{Spacer} by comparing its theses against initial ideations of existing research papers and outputs from existing SOTA LLMs.
To get a representative sample of high-quality human research without data contamination, we selected the abstracts of papers published in Nature, Science, Cell, Nature Methods, and Neuron between June 1st, 2025 and July 16th, 2025, all outside the knowledge cutoff of \textit{Spacer} and its training data.
These abstracts were then converted into thesis paragraphs as in the beginning of \cref{sec:experiments}.
For comparison, we prompted GPT-5, Gemini 2.5 Pro, Claude Opus 4, DeepSeek-R1-0528, and Grok 4 to create a paragraph describing an innovative research idea.
We randomly sampled 52 thesis paragraphs from each of the 7 classes (\textit{Spacer}, published papers, and 5 LLMs in total), resulting in a total of 364 theses analyzed.

To compare these classes easily, we chose to analyze text embeddings of the resulting thesis paragraphs.
The model used for embedding was Qwen3-Embedding-8B  \cite{qwen3embedding}, chosen for its strong performance on text embedding tasks.
Since textual embeddings include information about style  \cite{icard2025embeddingstyletopicsanalyzing}, we further normalized stylistic properties in each thesis paragraph with Claude Sonnet 4 and Kimi K2 before embedding.
These two preprocessing LLMs were deliberately chosen to be distinct from the LLMs to ensure neutrality. 
The full prompts used to generate and process these paragraphs, as well as the paragraphs themselves, can be found in Appendix~\ref{app:embeddingprompt} and our \href{https://github.com/Asteromorph-Corp/Spacer}{Github Repository}.

\begin{figure}[p]
    \centering
    \includegraphics[width=0.98\textwidth]{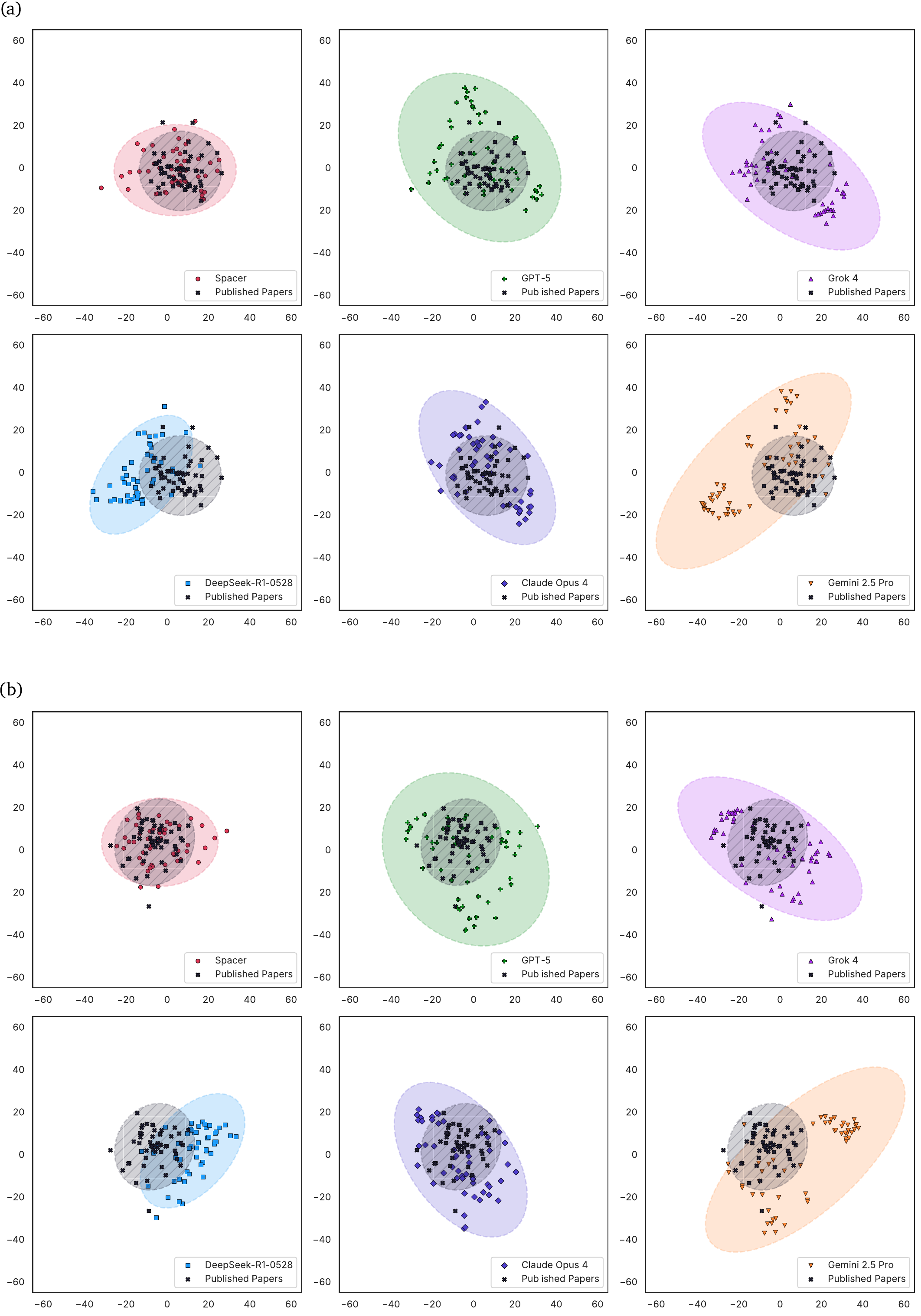}
    \caption{PCA results of 364 research theses from \textit{Spacer}, 5 SOTA LLMs, and ideations of published papers. Each pane displays a single generated class with the published class to emphasize their difference. Processing models used are (a) Claude Sonnet 4. (b) Kimi K2.}
    \label{fig:embedding_PCA}
\end{figure}

We first applied a principal component analysis (PCA) on the entire set of 364 vectors, and visualized each of the 6 generated classes alongside the published papers in \cref{fig:embedding_PCA}.
Our plots clearly show \textit{Spacer}'s outputs are the closest to the published samples.
Although LLMs are also capable of generating outputs that are close, they display high variance, with many samples semantically distant from the human ones.
This result is consistent across the two processing models used.

\begin{figure}[ht]
    \centering
    \includegraphics[width=0.98\textwidth]{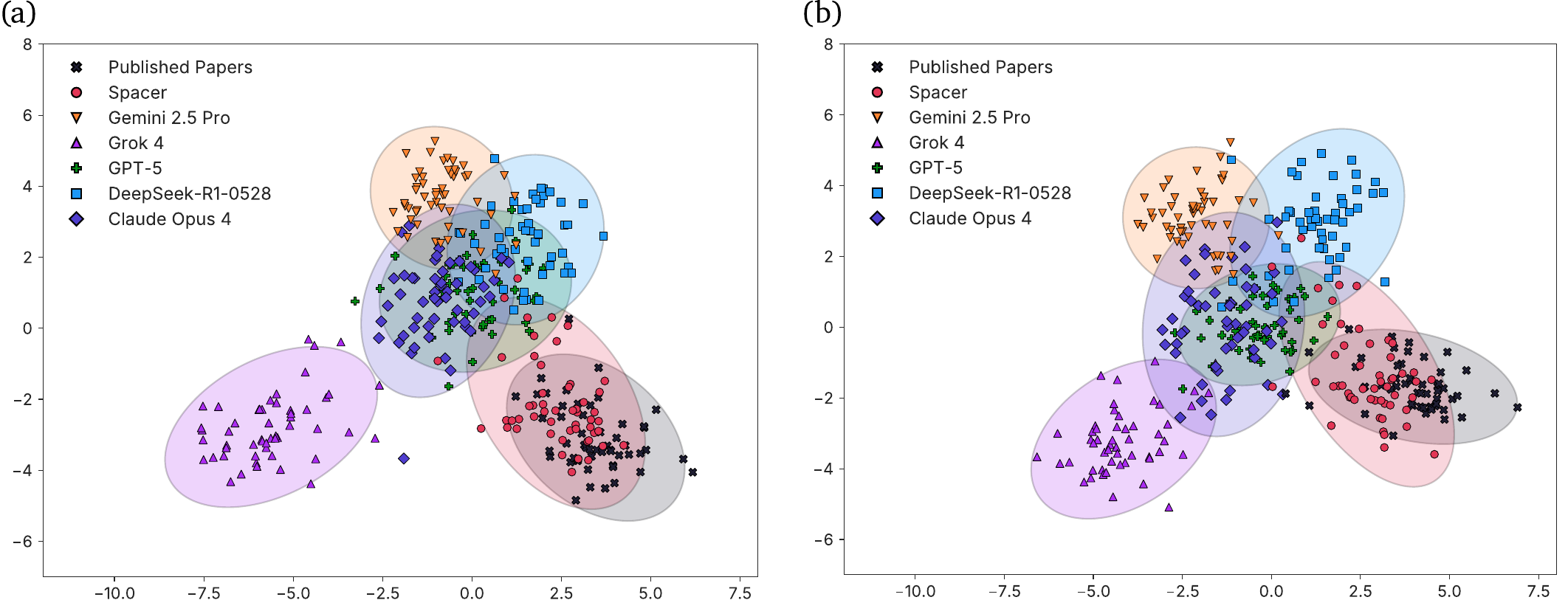}
    \caption{LDA results of 364 research theses from \textit{Spacer}, 5 SOTA LLMs, and ideations of published papers. Full embeddings were reduced to 128-dimensional vectors with PCA before the LDA step. Processing models used are (a) Claude Sonnet 4. (b) Kimi K2.}
    \label{fig:embedding_LDA}
\end{figure}

We next performed an linear discriminant analysis (LDA), with an initial PCA step to compress inputs into 128-dimensional vectors.
The results were projected onto the first two axes and then plotted in \cref{fig:embedding_LDA}.
We observe that LLM proposals are largely discriminated from published papers, while \textit{Spacer} outputs overlap closely. 
Again, the results are consistent between the processing models, demonstrating that theses from \textit{Spacer} are much more semantically aligned with those of expert research than those from any other SOTA LLM.

\begin{figure}[ht]
    \centering
    \includegraphics[width=0.98\textwidth]{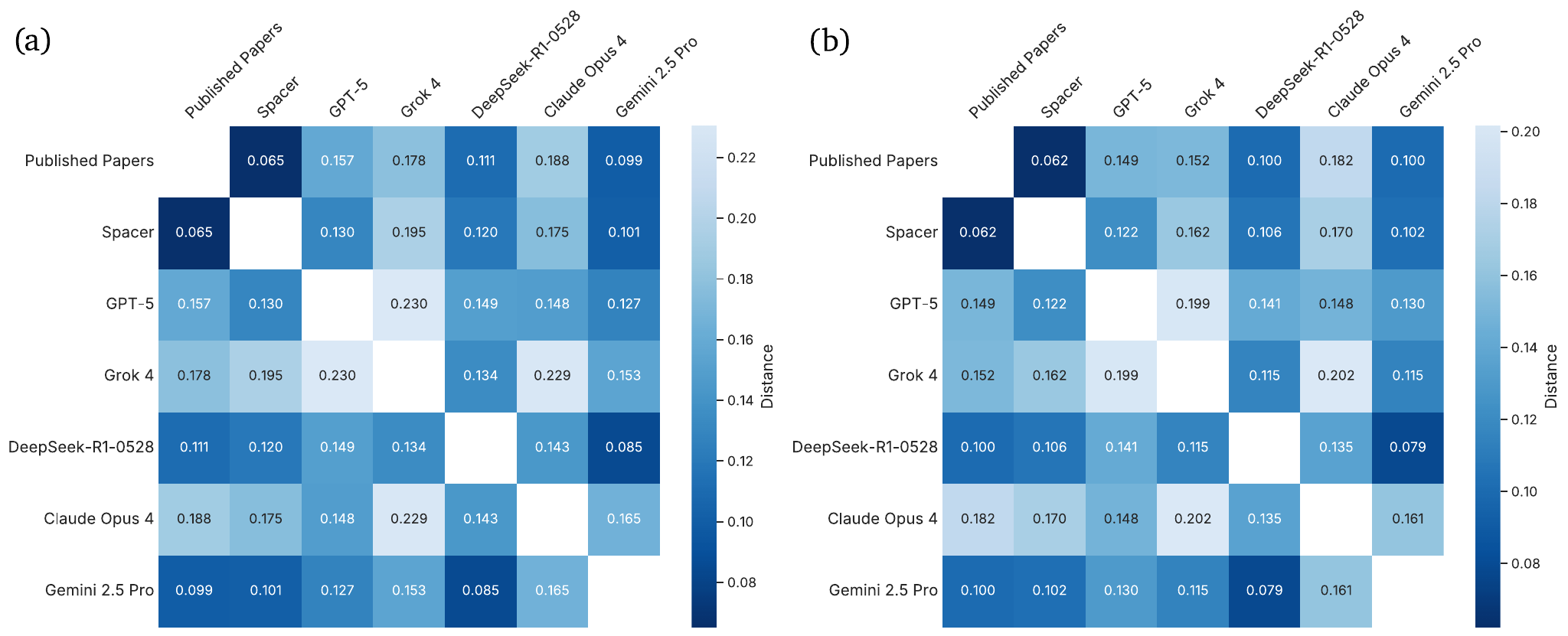}
    \caption{Heatmap of energy distances between \textit{Spacer}, 5 SOTA LLMs, and ideations of published papers. Distances were computed with full embedding vectors. Processing models used are (a) Claude Sonnet 4. (b) Kimi K2.}
    \label{fig:embedding_heatmap}
\end{figure}

Figure~\ref{fig:embedding_heatmap} shows the energy distances between the full vectors of each class, given by the formula below:
\begin{equation}\label{eq:energy_distance}
    \hat{D}_E(X,Y) = \frac{2}{mn}\sum_{i=1}^{m}\sum_{j=1}^{n}\left\| x_i-y_j \right\|_2 - \frac{1}{m^{2}}\sum_{i=1}^{m}\sum_{k=1}^{m}\left\| x_i-x_k \right\|_2 - \frac{1}{n^{2}}\sum_{j=1}^{n}\sum_{l=1}^{n}\left\| y_j-y_l \right\|_2
\end{equation}
We did not use KL-divergence between each class since we could not perform a normality test on the classes (each vector was 4096-dimensional while we only had 52 samples for each class).
This quantitative analysis again shows \textit{Spacer} is the closest class to the published papers; in fact, they have the smallest distance out of all pairs in our sample.

Overall, \textit{Spacer} is significantly closer to top human research than SOTA LLMs, regardless of the choice of metric.
This analysis suggests that our architectural modifications are a key augmentation without which current LLMs cannot reliably generate research concepts at the expert level.


\clearpage
\section{Discussion}

In this report, we have introduced \textit{Spacer}.
\textit{Spacer} is the first framework to acknowledge the creative limitations of AI models stemming from their reliance on training data and thus adopt an alternate approach---\textbf{deliberate decontextualization}.
This approach, not relying on serendipity, is closely aligned to our ultimate goal of achieving \textbf{engineered scientific inspiration}.
We take Keyword Sets as our central medium, and postulate that new concepts can be discovered by searching for connections in them.
With this postulate, we generate reasonable yet novel Keyword Sets with \textsc{Nuri} and guide LLM-based frameworks to explore their internal links.
Our experiments on reconstructing past research ideas from Keyword Sets have demonstrated that research inspirations indeed lie within these sets.
Moreover, we find \textit{Spacer} to be more semantically aligned with human research than those from other SOTA LLMs, further confirming our core thesis.

\paragraph*{Why does `inspiration' matter?}

For most of history, humanity has regarded itself as the only entity capable of conceiving new ideas and realizing them.
The rise of AI has eroded many similar beliefs about human uniqueness, and we believe that the next step for AI will be to push the limits in humanity's pursuit of scientific discovery.
However, we remain skeptical of current LLM-based approaches, and more generally, of approaches relying solely on generative models based on pretraining and prompting.
Such approaches model human inspirations primarily through randomness, which may result in mode-collapsed ideas.

From this perspective, \textit{Spacer} is the closest system to automating the expansion of humanity's knowledge.
A complete shift to such a paradigm would empower humanity to advance civilization at an unprecedented speed, and \textit{Spacer} is our first step in this direction.

\paragraph*{Cost.}

Currently, \textit{Spacer} costs less than \$3 to generate a single Statement.
Compared to the expected value of a potentially significant discovery, this is an attractive cost.

Most of the cost of the \textit{Spacer} pipeline comes from LLM inference; the cost of generating a Keyword Set from \textsc{Nuri} is effectively zero, giving room to enhance \textsc{Nuri} with more compute.
As \textit{Spacer} is agnostic to the transformer architecture that powers most current LLMs, more efficient language modeling techniques can be directly integrated into our pipeline, whether they are further extensions of transformers or more fundamental innovations.

\paragraph*{Limitations.}

Human inspiration has no single source: inspection of extensive data, experience in a different field \cite{gilmer2022constantlowerboundunionclosed}, and daydreams \cite{schultz1890} all have provided the spark for monumental discoveries.
Thus, we cannot assert that deliberate decontextualization and search in Keyword Sets is the single most suitable, or the most human-like, way to synthesize scientific inspiration.
We are much more confident in our work's optimality for LLM-driven methods; beyond this paradigm, other approaches may prove equally or more effective.

Even within the LLM-driven space, \textit{Spacer} can be augmented in several different ways.
Improvements in the base models, as well as reinforcement learning approaches to fine-tune each component, will enhance \textit{Spacer}'s performance.
On a more fundamental level, we conjecture that non-autoregressive language models will have a greater capacity for creative thought.

\paragraph*{Future Directions.}

Not all scientific inspirations bear fruit; experiments must be designed and performed to validate their hypotheses before they can become new knowledge or be applied to engineering.
Therefore, this materialization step is a hard requirement for any attempt at full automation of the human scientific process.
In the future, we will extend \textit{Spacer} to further flesh out its accepted Statement into an executable research plan, preferably by in-silico experiments.

As a preliminary experiment, we examined the abilities of current LLMs to perform this concretization.

For this task, we chose Grok 4 for its PhD-level capabilities \cite{xai2025grok4}, and instructed it to formulate a validation plan based on the Statements generated by \textit{Spacer}.
The proposed plan contained all information necessary to conduct the research and only required minor revisions before immediate implementation.
The complete outputs are provided in Appendix~\ref{sec:grokresult}.

This result demonstrates that automated knowledge expansion through \textit{Spacer} is already within reach.
This will be our next step: developing an end-to-end pipeline that carries out the entire research process without any human intervention.
We plan to apply robotics and \textit{in silico} tools to conduct the proposed experiments automatically, while also devising more sophisticated methods for research plan construction.

Although this work is focused mainly on biological research, there are no obstructions to implementing \textit{Spacer} in any field of science, or indeed, any field where human inspiration is of importance.
In fact, we expect \textit{Spacer}'s approach to be most useful in fields where the impacts of ideas are heavy-tailed---i.e., where creative breakthroughs have an outsized effect.
Straightforward future extensions include physics, machine learning, and economics.
However, \textit{Spacer} need not be constrained to a single domain; ultimately, \textsc{Nuri}'s graph could be unified to explore arbitrary areas of science, finding truly deep, creative, and interconnected scientific concepts.

\paragraph*{Conclusion.}

\textit{Spacer} is a scientific discovery system that automates academic research by capturing the essence of humanity's inspiration in science.
We have overcome the creative limitations of LLMs by using \textsc{Nuri} to search for unexplored connections between decontextualized information---keywords.
Through this critical augmentation, \textit{Spacer} has emergently conceived original concepts, something long thought possible only for humans.

Researchers are now actively pursuing the creation of artificial superintelligence (ASI), a system capable of surpassing human knowledge and creativity.
Contributing to this ambition, \textit{Spacer} has demonstrated the potential to engineer scientific inspiration.
With further refinement and scaffolding tailored to biology, it has come tantalizingly close to actual scientific progress.
As we advance towards automating the expansion of knowledge, we believe our work marks a pivotal step forward in this collective journey.

\clearpage
\section{Technical Details}


\subsection{System Designs\label{subsec:frameworks}}

The primary language we used for development is Rust, chosen for its type safety and strong compile-time guarantees.
This choice minimized the risk of runtime errors and ensured that our code is robust and maintainable.
We also built a custom framework that implements agentic AI in Rust, as existing solutions did not meet our needs.

Multiple tasks we encountered during the research consist of several subtasks.
While each subtask was well-defined, it could yield different numbers of outputs depending on the input.
This variation caused complications when applying traditional approaches for handling multi-step process.
As such, we developed an incremental scheduling framework to maximize the parallelism between subtasks.

Another task that played a crucial role throughout the various steps of \textit{Spacer} was retrieving relevant articles to feed into the LLM.
To address this, we created a search database for scientific paper abstracts using data acquired from OpenAlex \cite{openalexfullyopenindexscholarly}.
The database utilizes custom filters and LLM-based post-processing to produce the most accurate search results.


\subsection{Model Specifics}

\textit{Spacer} utilizes proprietary SOTA LLMs such as o3 \cite{openai2025o3o4mini}, Grok 4 \cite{xai2025grok4}, Gemini 2.5 Pro \cite{deepmind2025gemini25}, and Claude Opus 4 \cite{anthropic2025claude4} as backbones for its components.
It also employs fine-tuned open-weight models such as DeepSeek-R1 \cite{deepseekai2025deepseekr1incentivizingreasoningcapability}, and Gemma 3 \cite{gemmateam2025gemma3technicalreport}.
The knowledge cutoffs for all models are specified in Table~\ref{tab:knowledgecutoff}.

As a side effect of reinforcement learning, the proprietary models lack variety \cite{yue2025}.
These tendencies significantly impair the creative capabilities of \textit{Spacer}.
Hence, we fine-tuned the LLMs to realign them away from these inclinations.

For instance, \textsc{Weaver} and \textsc{Sketcher} were fine-tuned for research initiative structuralization and research goal suggestion respectively.
We used DeepSeek-R1 for \textsc{Weaver} and Gemma 3 27B for \textsc{Sketcher}.
We fine-tuned \textsc{Weaver} using a custom dataset comprising pairs of Keyword Sets and their corresponding inspirational ideas.
Figure~\ref{fig:weaverdataexample} in Appendix~\ref{subsec:data_specs} shows an example entry from the training dataset.

\begin{table}[htb]
    \centering
    \caption{Knowledge cutoffs of models used in \textit{Spacer}.}
    \begin{tabular}{cc}
        \toprule
        Model                                                                       & Knowledge Cutoff \\
        \midrule
        Grok 4 \cite{xai2025grok4}                                                  & November 2024    \\
        Gemini 2.5 Flash/Pro \cite{deepmind2025gemini25}                            & January 2025     \\
        o3 \cite{openai2025o3o4mini}                                                & June 2024        \\
        GPT-4.1/GPT-4.1 mini \cite{openai2025gpt41}                                 & June 2024        \\
        Claude Opus 4 \cite{anthropic2025claude4}                                   & November 2024    \\
        Claude Sonnet 4 \cite{anthropic2025claude4}                                 & March 2025       \\
        \midrule
        DeepSeek-R1 \cite{deepseekai2025deepseekr1incentivizingreasoningcapability} & January 2025     \\
        Gemma 3 4B/27B \cite{gemmateam2025gemma3technicalreport}                    & March 2025       \\
        \bottomrule
    \end{tabular}
    \label{tab:knowledgecutoff}
\end{table}


\subsection{Hardwares\label{subsec:hardware_specs}}

We used a single node of 8\,$\times$\,NVIDIA H100 GPUs for training and serving the LLMs, and another node with an Intel i9-14900F CPU, 192GB of RAM, and a RTX 4090 GPU for the rest of the components.

\clearpage
{\hypersetup{urlcolor=black}
    \printbibliography[title=References]
}

\clearpage
\appendix
\section*{Contributions and Acknowledgments\label{sec:contributions}}

\begin{multicols}{2}
    \color{DodgerBlue4} \textbf{Research \& Engineering} \\
    Donghyun Koh\textsuperscript{*} \\
    Geonho Nah\textsuperscript{*} \\
    Haneul Choi \\
    Hojin Yoo \\
    Jaeyeong Kim \\
    Jiho Park \\
    Johyun Park \\
    Juneau Jung \\
    Minhyeong Lee\textsuperscript{*, \textdagger} \\

    \columnbreak

    Seongjun Kim \\
    Seunghyun Moon\textsuperscript{*} \\
    Seungwon Kim \\
    Sungbin Moon \\
    Suyoung Hwang\textsuperscript{*} \\
    Taehoon Hwang \\
    Youngjun Cho\textsuperscript{*}

    \color{Green4} \textbf{Business \& Operation} \\
    Seongjin Seol \\
    Yeji Kim
\end{multicols}
\noindent Authors are listed alphabetically within each team. \\
\small
\noindent \textsuperscript{*} Main contributors. \ \textsuperscript{\textdagger} Correspondence to Minhyeong Lee (\url{mh.lee@asteromorph.com})
\normalsize

\section*{Appendix}

\renewcommand{\thesubsection}{\Alph{subsection}}
\renewcommand{\thesubsubsection}{\thesubsection\arabic{subsubsection}}
\setcounter{subsection}{0}


\subsection{Prompts}


\subsubsection{Prompts for the Revealing Framework}

\begin{center}
    \begin{tcolorbox}[title={\textbf{Simplified Prompt for the Weaver}}, colback=bluebg, colframe=blueheader]
        \begin{lstlisting}[breaklines=true, basicstyle=\ttfamily\footnotesize, columns=fullflexible, breakindent=0pt, frame=none, breakatwhitespace=true]
[System Prompt]
Role:
You are a domain expert tasked with generating a novel research hypothesis from a compact set of biologically and technically relevant keywords.

[User Prompt]
Goal:
You must identify a coherent subset of these terms, discard incompatible ones, and formulate a single logically unified research concept.

Instructions:
Each keyword denotes an experimental tool, biological entity, or conceptual mechanism.
Avoid speculating about results or impacts.

Steps:
- Define each term's technical role.
- Identify logical constraints and compatibilities.
- Construct a conceptual framework.
- Write a self-contained paragraph describing the research idea.
\end{lstlisting}
    \end{tcolorbox}
    \captionof{figure}{Simplified prompt for \textsc{Weaver}.}\label{fig:chefprompt}
\end{center}

\begin{center}
    \begin{tcolorbox}[title={\textbf{Simplified Prompt for the Sketcher}}, colback=bluebg, colframe=blueheader]
        \begin{lstlisting}[breaklines=true, basicstyle=\ttfamily\footnotesize, columns=fullflexible, breakindent=0pt, frame=none, breakatwhitespace=true]
[System Prompt]
Role:
You are a domain specialist who distills visionary yet precise scientific objectives from a concise list of technical keywords.

[User Prompt]
Goal:
Craft a sentence that states an ambitious scientific breakthrough attainable within 15-30 years.

Instructions:
- Derive the idea by uncovering conceptual links among the provided terms.
- The objective must be transformative, measurable, and scientifically plausible.
- Do not repeat the original keywords; translate their concepts.
\end{lstlisting}
    \end{tcolorbox}
    \captionof{figure}{Simplified prompt for \textsc{Sketcher}.}\label{fig:sketcherprompt}
\end{center}

\begin{center}
    \begin{tcolorbox}[title={\textbf{Simplified Prompt for the Untrained LLM Generating the Thesis}}, colback=bluebg, colframe=blueheader]
        \begin{lstlisting}[breaklines=true, basicstyle=\ttfamily\footnotesize, columns=fullflexible, breakindent=0pt, frame=none, breakatwhitespace=true]
[System Prompt]
Role:
You are a scientific strategist who converts broad research ambitions into tightly scoped, methodologically sound study proposals.

[User Prompt]
Goal:
Identify one precise sub-problem that advances the ultimate objective and craft a single, rigorously grounded research idea.

Instructions:
- Choose a sub-problem directly linked to the goal.
- Offer one novel mechanistic insight or framework rooted in established science.
- Express the idea in ~100 academic words, as one paragraph.
- Exclude detailed protocols, speculative mechanisms, or exaggerated claims.
\end{lstlisting}
    \end{tcolorbox}
    \captionof{figure}{Simplified prompt for the untrained LLM generating the Thesis.}\label{fig:untrainedprompt}
\end{center}


\newpage
\subsubsection{Prompts for the Scaffolding Framework}

\begin{center}
    \begin{tcolorbox}[title={\textbf{Simplified Prompt for Graph Generation}}, colback=bluebg, colframe=blueheader]
        \begin{lstlisting}[breaklines=true, basicstyle=\ttfamily\footnotesize, columns=fullflexible, breakindent=0pt, frame=none, breakatwhitespace=true]
[System Prompt]
Role:
You are a logic analyst tasked with converting a research idea into a structured reasoning graph.

[User Prompt]
Task:
Verify whether the provided rationales logically support the main research concept, identify any missing links, and generate necessary intermediate conclusions.

Steps:
- Use all provided rationales exactly as given.
- Form intermediate nodes that bridge groups of rationales toward the main concept.
- Organize the graph as a tree: rationale -> intermediate -> main concept.
\end{lstlisting}
    \end{tcolorbox}
    \captionof{figure}{Simplified prompt for the Graph Generation phase in the Scaffolding Framework.}\label{fig:sfprompt}
\end{center}

\begin{center}
    \begin{tcolorbox}[title={\textbf{Simplified Prompt for Rationale Validation}}, colback=bluebg, colframe=blueheader]
        \begin{lstlisting}[breaklines=true, basicstyle=\ttfamily\footnotesize, columns=fullflexible, breakindent=0pt, frame=none, breakatwhitespace=true]
[System Prompt]
Role:
You are a scientific evaluator responsible for verifying the scientific correctness of a specific proposition.

[User Prompt]
Task:
Assess whether the claim aligns with established scientific understanding by searching for evidence from peer-reviewed literature using the article search tool.

Procedure:
- Use the article search tool with precision using the exact terms from the proposition.
- Classify each source's relevance to the proposition.
- Return a JSON object that contains:
  - the proposition text
  - your final evaluation status
  - the list of DOIs that support your assessment
\end{lstlisting}
    \end{tcolorbox}
    \captionof{figure}{Simplified prompt for the Rationale Validation phase in the Scaffolding Framework.}
    \label{fig:sfvalidationprompt}
\end{center}


\subsubsection{Prompts for the Assessment Framework}

\begin{center}
    \begin{tcolorbox}[title={\textbf{Simplified Prompt for Exploratory Analysis}}, colback=bluebg, colframe=blueheader]
        \begin{lstlisting}[breaklines=true, basicstyle=\ttfamily\footnotesize, columns=fullflexible, breakindent=0pt, frame=none, breakatwhitespace=true]
[System Prompt]
Role:
You are a multidisciplinary scientist who gives balanced, constructive critiques of research proposals.

[User Prompt]
Task:
Critically evaluate the proposal and provide a two-part review of validity and irrationality.

Steps:
- Read the proposal's `concept' and `rationale' exactly as given.
- Comprehensively evaluate the proposal from a scientific perspective.
- Write a 3-4 sentence overall summary.
- List well-supported points under validity.
- List inconsistent or unreasonable points under irrationality.
- Skip trivial editing remarks.

Input: {{statement}}

Output (JSON):
{
  "summary": "...",
  "validity": ["...", ...],
  "irrationality": ["...", ...]
}
\end{lstlisting}
    \end{tcolorbox}
    \captionof{figure}{Simplified prompt for the reviewer in the Assessment Framework.}\label{fig:reviewerprompt}
\end{center}

\begin{center}
    \begin{tcolorbox}[title={\textbf{Simplified Prompt for Specified Inspection}}, colback=bluebg, colframe=blueheader]
        \begin{lstlisting}[breaklines=true, basicstyle=\ttfamily\footnotesize, columns=fullflexible, breakindent=0pt, frame=none, breakatwhitespace=true]
[System Prompt]
Role:
You are a meta-reviewer who rates the seriousness of each irrationality flagged by reviewers.

[User Prompt]
Task:
Score every irrationality on severity and explain your decision.

Steps:
- Focus only on methodological flaws, feasibility issues, and scientific impossibilities.
- Choose one option per irrationality:
  A Fatal  B Serious  C Moderate  D Minor  E Negligible.
- Give 1-2 sentences of rationale for each score.

Input:
Research Idea:
{{statement}}

Irrationality:
{{irrationality}}

Output (JSON):
{
  "meta_review": [
    { "option": "A|B|C|D|E", "rationale": "..." },
    ...
  ]
}
\end{lstlisting}
    \end{tcolorbox}
    \captionof{figure}{Simplified prompt for the judge in the Assessment Framework.}\label{fig:afjudgeprompt}
\end{center}


\subsubsection{Full Prompt for Embedding Dataset Generation\label{app:embeddingprompt}}

\begin{center}
    \begin{tcolorbox}[title={\textbf{Full Prompt for Research Idea Generation of SOTA LLMs}}, colback=bluebg, colframe=blueheader]
        \begin{lstlisting}[breaklines=true, basicstyle=\ttfamily\footnotesize, columns=fullflexible, breakindent=0pt, frame=none, breakatwhitespace=true]
[System Prompt]
You are a creative biomedical researcher proposing innovative research directions.

[User Prompt]
Generate an innovative biomedical research concept addressing an unresolved biological question.

## Format Guidelines
- Length: 100-150 words
- Structure:
  - Present the innovative approach or methodology
  - Link the approach to testable predictions
  - Explain how this strategy addresses the mechanistic question
- Voice: First-person plural ("we")
- Tense: Present for established facts, conditional/future for proposed work
- Focus: Mechanistic reasoning over technical details

## Output
Research proposal paragraph only.
\end{lstlisting}
    \end{tcolorbox}
    \captionof{figure}{Full prompt for research idea generation of SOTA LLMs.}
    \label{fig:zeroshotllmprompt}
\end{center}

\begin{center}
    \begin{tcolorbox}[title={\textbf{Full Prompt for Converting a Statement into an Unstructured Paragraph}}, colback=bluebg, colframe=blueheader]
        \begin{lstlisting}[breaklines=true, basicstyle=\ttfamily\footnotesize, columns=fullflexible, breakindent=0pt, frame=none, breakatwhitespace=true]
[System Prompt]
You are a scientific writer who synthesizes structured research concepts into cohesive scientific narratives.

[User Prompt]
Transform the JSON research proposal into a unified paragraph.

## Input Structure
- "concept": Core hypothesis
- "rationale": Supporting evidence array

## Synthesis Guidelines
1. Lead with the innovative approach from "concept"
2. Integrate "rationale" as supporting logic
3. Maintain mechanistic focus throughout
4. Conclude with testable predictions or implications
5. Ensure each statement flows logically from the previous one, creating a coherent narrative chain

## Format Guidelines
- Length: 100-150 words
- Structure:
  - Present the innovative approach or methodology
  - Link the approach to testable predictions
  - Explain how this strategy addresses the mechanistic question
- Voice: First-person plural ("we")
- Tense: Present for established facts, conditional/future for proposed work
- Focus: Mechanistic reasoning over technical details

## Input
{{statement}}

## Output
Research proposal paragraph only.
\end{lstlisting}
    \end{tcolorbox}
    \captionof{figure}{Full prompt for converting a Statement into an unstructured paragraph.}
    \label{fig:structuredtoparagraphprompt}
\end{center}

\begin{center}
    \begin{tcolorbox}[title={\textbf{Full Prompt for Converting Published Reserach Abstract into Paragraph Form}}, colback=bluebg, colframe=blueheader]
        \begin{lstlisting}[breaklines=true, basicstyle=\ttfamily\footnotesize, columns=fullflexible, breakindent=0pt, frame=none, breakatwhitespace=true]
[System Prompt]
You are a scientific writer who transforms research findings into forward-looking research proposals with clarity and precision.

[User Prompt]
Transform the following research abstract into a research proposal paragraph.

## Core Requirements
1. Identify the mechanistic question or knowledge gap
2. Extract the novel hypothesis or innovative approach
3. Convert specific results into testable predictions
4. Preserve key scientific terminology

## Format Guidelines
- Length: 100-150 words
- Structure:
  - Present the innovative approach or methodology
  - Link the approach to testable predictions
  - Explain how this strategy addresses the mechanistic question
- Voice: First-person plural ("we")
- Tense: Present for established facts, conditional/future for proposed work
- Focus: Mechanistic reasoning over technical details

## Input
Abstract: {{abstract_text}}

## Output
Research proposal paragraph only.
\end{lstlisting}
    \end{tcolorbox}
    \captionof{figure}{Full prompt for converting published research abstracts into paragraph form.}
    \label{fig:dressingextractprompt}
\end{center}

\begin{center}
    \begin{tcolorbox}[title={\textbf{Full Prompt for Rephrasing and Summarizing Research Thesis Paragraphs}}, colback=bluebg, colframe=blueheader]
        \begin{lstlisting}[breaklines=true, basicstyle=\ttfamily\footnotesize, columns=fullflexible, breakindent=0pt, frame=none, breakatwhitespace=true]
[System Prompt]
You are a scientific editor who distills research ideas into concise statements capturing their unique insight, approach, and distinguishing feature.

[User Prompt]
## Task
Reformulate the following research idea into a concise statement focusing on the core insight, conceptual approach, and unique distinguishing element.

## Requirements
- Length: 45-55 words total
- Focus ONLY on:
  1. The central mechanistic insight or hypothesis
  2. The conceptual strategy to address it
  3. The unique angle that distinguishes this research
- Remove unnecessary details, background context, and procedural specifics
- Preserve essential scientific specificity

## Writing Style
- Integrate elements into a flowing statement
- Use active, direct language
- Eliminate qualifying phrases
- Disregard the original phrasing and structure; reformulate the idea entirely in your own distinctive voice

## Input
{{idea_paragraph}}

## Output
The reformulated statement only.
\end{lstlisting}
    \end{tcolorbox}
    \captionof{figure}{Full prompt for rephrasing and summarizing research thesis paragraphs.}
    \label{fig:dressingsummarizerprompt}
\end{center}


\subsubsection{Full Prompt for the Judge Evaluating the Reconstructed Papers\label{app:judgeprompt}}

\begin{center}
    \begin{tcolorbox}[title={\textbf{Full Prompt for the Judge Evaluating the Reconstructed Papers (Logic)}}, colback=bluebg, colframe=blueheader]
        \begin{lstlisting}[breaklines=true, basicstyle=\ttfamily\footnotesize, columns=fullflexible, breakindent=0pt, frame=none, breakatwhitespace=true]
[System Prompt]

[User Prompt]
## Task
Identify whether the provided ideas share the same logical structure or not.

## Elements to Consider
If the ideas share the logical structure, they may be conducted in the same way.

## Instructions
1. Read the two provided ideas carefully.
2. Only return "yes" or "no" without any additional text or explanation.

## First Idea
{{first_idea}}

## Second Idea
{{second_idea}}
\end{lstlisting}
    \end{tcolorbox}
    \captionof{figure}{Full Prompt for the judge evaluating the reconstructed papers (Logic).}
    \label{fig:logicjudgeprompt}
\end{center}

\begin{center}
    \begin{tcolorbox}[title={\textbf{Full Prompt for the Judge Evaluating the Reconstructed Papers (Topic)}}, colback=bluebg, colframe=blueheader]
        \begin{lstlisting}[breaklines=true, basicstyle=\ttfamily\footnotesize, columns=fullflexible, breakindent=0pt, frame=none, breakatwhitespace=true]
[System Prompt]

[User Prompt]
## Task
Identify whether the provided ideas share the same topic or not.

## Elements to Consider
If the ideas share the same subject matter or theme, they are likely to be on the same topic.

## Instructions
1. Read the two provided ideas carefully.
2. Only return "yes" or "no" without any additional text or explanation.

## First Idea
{{first_idea}}

## Second Idea
{{second_idea}}
\end{lstlisting}
    \end{tcolorbox}
    \captionof{figure}{Full Prompt for the judge evaluating the reconstructed papers (Topic).}
    \label{fig:topicjudgeprompt}
\end{center}

\begin{center}
    \begin{tcolorbox}[title={\textbf{Full Prompt for the Judge Evaluating the Reconstructed Papers (Objective)}}, colback=bluebg, colframe=blueheader]
        \begin{lstlisting}[breaklines=true, basicstyle=\ttfamily\footnotesize, columns=fullflexible, breakindent=0pt, frame=none, breakatwhitespace=true]
[System Prompt]

[User Prompt]
## Task
Identify whether the provided ideas share the same objective or not.

## Elements to Consider
If the ideas share the same goal or purpose, they are likely to have the same objective.

## Instructions
1. Read the two provided ideas carefully.
2. Only return "yes" or "no" without any additional text or explanation.

## First Idea
{{first_idea}}

## Second Idea
{{second_idea}}
\end{lstlisting}
    \end{tcolorbox}
    \captionof{figure}{Full Prompt for the judge evaluating the reconstructed papers (Objective).}
    \label{fig:objectivejudgeprompt}
\end{center}

\begin{center}
    \begin{tcolorbox}[title={\textbf{Full Prompt for the Judge Evaluating the Reconstructed Papers (Approach)}}, colback=bluebg, colframe=blueheader]
        \begin{lstlisting}[breaklines=true, basicstyle=\ttfamily\footnotesize, columns=fullflexible, breakindent=0pt, frame=none, breakatwhitespace=true]
[System Prompt]

[User Prompt]
## Task
Identify whether the provided ideas share the same approach or not.

## Elements to Consider
If the ideas share the same method or strategy, they are likely to have the same approach.

## Instructions
1. Read the two provided ideas carefully.
2. Only return "yes" or "no" without any additional text or explanation.

## First Idea
{{first_idea}}

## Second Idea
{{second_idea}}
\end{lstlisting}
    \end{tcolorbox}
    \captionof{figure}{Full Prompt for the judge evaluating the reconstructed papers (Approach).}
    \label{fig:approachjudgeprompt}
\end{center}

\begin{center}
    \begin{tcolorbox}[title={\textbf{Full Prompt for the Judge Evaluating the Reconstructed Papers (Overall)}}, colback=bluebg, colframe=blueheader]
        \begin{lstlisting}[breaklines=true, basicstyle=\ttfamily\footnotesize, columns=fullflexible, breakindent=0pt, frame=none, breakatwhitespace=true]
[System Prompt]

[User Prompt]
## Task
Identify whether the provided ideas are identical or not.

## Elements to Consider
Identify if the ideas share:
- Biological target (e.g., protein, gene, pathway)
- Conceptual focus (e.g., mechanism, process)
- Experimental approach (e.g., technique, method)
- Unique elements (e.g., specific reagents, hardware)

## Instructions
1. Read the two provided ideas carefully.
2. Compare them based on the elements listed above.
3. Only return "yes" or "no" without any additional text or explanation.

## First Idea
{{first_idea}}

## Second Idea
{{second_idea}}
\end{lstlisting}
    \end{tcolorbox}
    \captionof{figure}{Full Prompt for the judge evaluating the reconstructed papers (Overall).}
    \label{fig:overalljudgeprompt}
\end{center}


\subsubsection{Prompt for Generating Experimental Protocol (Grok 4) \label{app:grok4prompt}}

\begin{center}
    \begin{tcolorbox}[title={\textbf{Simplified Prompt for Experimental Protocol Agent}}, colback=bluebg, colframe=blueheader]
        \begin{lstlisting}[breaklines=true, basicstyle=\ttfamily\footnotesize, columns=fullflexible, breakindent=0pt, frame=none, breakatwhitespace=true]
[System Prompt]
Role:
You are an experimental biologist who designs comprehensive laboratory protocols to validate research hypotheses. You excel at creating practical, step-by-step experimental strategies using established techniques and commercially available resources in standard molecular biology laboratories.

[User Prompt]
Task:
Transform a research hypothesis into a hierarchical experimental validation plan with multiple experimental branches that adapt based on outcomes.

Instructions:
- Analyze the hypothesis to identify critical experimental objectives
- Design 2-4 complementary base experiments that address the core questions
- For each experiment, develop subsequent steps based on probable outcomes
- Create branches only where scientifically necessary
- Ensure each step is measurable and directly actionable

Requirements:
- Use only established, commercially available techniques and reagents
- Include specific procedures, concentrations, and analysis methods
- List required cell lines, reagents, and equipment for each step
- Predict realistic outcomes based on biological precedent

Output Format:
1. Hypothesis summary
2. Root experiments containing:
   - Rationale and biological question addressed
   - Detailed procedures with specific parameters
   - Required resources (cell lines, reagents, equipment)
   - Expected outcomes and most probable result
   - Conditional branches for follow-up experiments
3. Integration strategy explaining how experiments relate
4. Contingency plans for alternative approaches

Constraints:
- Focus on practical feasibility over theoretical possibilities
- Base outcome predictions on established biological principles
- Consider technical limitations and experimental challenges
\end{lstlisting}
    \end{tcolorbox}
    \captionof{figure}{Simplified prompt for the \textsc{Grok 4}, experimental protocol design.}
    \label{fig:grok4exprompt}
\end{center}


\newpage
\subsection{Example of Experimental Protocol by Grok 4\label{sec:grokresult}}

\begin{center}
    \begin{tcolorbox}[breakable, title={ }, colback=blackbg, colframe=blackheader]
        \begin{noindent}
            \begin{lstlisting}[breaklines=true, basicstyle=\ttfamily\scriptsize, columns=fullflexible, breakindent=0pt, frame=none, breakatwhitespace=true, mathescape=true]
## HYPOTHESIS SUMMARY

This experimental design tests whether controlled injection of non-Gaussian noise into calcium signaling pathways via aperiodic modulation of extracellular calcium concentrations can restore oscillatory coherence in hepatocellular carcinoma (HCC) cells through stochastic resonance mechanisms, thereby re-engaging calcium-dependent cell cycle checkpoint regulation disrupted in malignancy. The rationale is based on observed disruptions in calcium homeostasis and oscillations in HCC cells compared to normal hepatocytes, with noise injection potentially enhancing signal coherence to suppress malignant proliferation selectively.

## STEP 1: Characterize Baseline Calcium Oscillation Patterns in Normal and HCC Hepatocytes

### Approach

**Rationale:** Establishing baseline differences in calcium signaling dynamics between normal and HCC cells is essential to confirm the hypothesis foundation, as disrupted oscillations in cancer cells are a prerequisite for testing restoration via noise. This step uses live-cell imaging, a feasible and established technique, to quantify oscillation frequency, amplitude, and coherence, addressing whether HCC cells exhibit less coherent patterns than normal cells.

**Procedure:**

1. Seed primary human hepatocytes (normal) and HepG2 HCC cells at $\mathrm{5x10^4}$ cells/well in 96-well plates coated with collagen I, culture in DMEM with 10% FBS at $\mathrm{37^{\circ}C}$, 5% CO2 for 24h.
2. Load cells with $\mathrm{5 \mu M}$ Fluo-4 AM calcium indicator dye for 30min at $\mathrm{37^{\circ}C}$, wash twice with HBSS (1.25mM CaCl2).
3. Perform live-cell confocal microscopy using a spinning-disk confocal system at $\mathrm{37^{\circ}C}$, acquiring images every 5s for 30min to capture spontaneous oscillations.
4. Stimulate with 100nM ATP to induce oscillations if spontaneous activity is low.
5. Analyze traces using ImageJ: quantify oscillation frequency (peaks/min), amplitude (fold-change from baseline), and coherence (autocorrelation function decay time constant, tau; lower tau indicates less coherence).
6. Include negative control (dye only, no cells) and positive control (ionomycin $\mathrm{1 \mu M}$ for maximal calcium response). Replicate in 3 independent experiments, n=50 cells/group per experiment, analyze with unpaired t-tests (p<0.05 significance).

**Required Resources:**

- Cell lines/models: Primary human hepatocytes (e.g., from Lonza) for normal physiology; HepG2 cells (ATCC) as HCC model, chosen for their well-characterized calcium dysregulation.
- Key reagents: Fluo-4 AM (Invitrogen, $\mathrm{5 \mu M}$); ATP (Sigma, 100nM); Ionomycin (Sigma, $\mathrm{1 \mu M}$); HBSS with 1.25mM CaCl2.
- Equipment: Spinning-disk confocal microscope (e.g., PerkinElmer UltraVIEW); 96-well plates; ImageJ software for analysis.

### Expected Outcomes

- Normal hepatocytes show coherent oscillations (tau >10min, frequency 0.1-0.5Hz, amplitude 2-5 fold).
- HCC cells show disrupted patterns (tau <5min, irregular frequency, sustained high baseline).
- No differences between cell types (null result).
- Technical artifacts (e.g., photobleaching leading to apparent low coherence).

### Most Probable Outcome

HCC cells exhibit disrupted oscillations with lower coherence (tau ~2-4min) and higher baseline calcium compared to normal hepatocytes (tau ~15min), based on established literature showing altered calcium homeostasis in cancer cells due to pump dysregulation; however, complete absence of oscillations is unlikely, as HCC cells retain some responsiveness, potentially yielding partial differences rather than stark contrasts.

### BRANCH 1.1: Validate Calcium Dysregulation Mechanisms

**Trigger:** Confirmation of disrupted oscillations in HCC cells (lower tau and irregular patterns).
**Approach:** Assess expression and function of key calcium regulators.

1. Perform qRT-PCR on RNA from both cell types using primers for PMCA1/4 (plasma membrane Ca2+-ATPase), TRPV channels, and housekeeping gene GAPDH; normalize to normal cells.
2. Western blot for PMCA proteins using anti-PMCA antibodies (Abcam, 1:1000 dilution, with beta-actin loading control).
3. Functional assay: Treat cells with PMCA inhibitor carboxyeosin (10  $\mathrm{\mu M}$) and repeat calcium imaging to confirm exacerbation of dysregulation.
4. Analyze with ANOVA, n=3 replicates.
   **Resources:** qRT-PCR kit (Qiagen); anti-PMCA antibodies (Abcam); carboxyeosin (Sigma, 10  $\mathrm{\mu M}$); immunoblotting system.
   **Expected Outcomes:** Lower PMCA expression/function in HCC; no differences; unexpected upregulation.
   **Most Probable:** Modest downregulation of PMCA in HCC (~50% reduction), consistent with known cancer adaptations for sustained signaling, but not complete loss, as cells maintain viability.

## STEP 2: Implement Controlled Noise Injection and Assess Oscillation Coherence Restoration

### Approach

**Rationale:** This core step tests the stochastic resonance hypothesis by introducing non-Gaussian noise via microfluidic modulation of extracellular calcium, measuring if it enhances coherence in HCC cells without affecting normal cells, using established microfluidic and imaging techniques for precise control and quantification.

**Procedure:**

1. Fabricate microfluidic device with perfusion channels (e.g., using soft lithography with PDMS) for rapid switching between solutions.
2. Seed HepG2 and normal hepatocytes in device chambers, load with Fluo-4 AM as in Step 1.
3. Perfuse with HBSS at 1.25mM CaCl2 baseline, then apply aperiodic modulation: switch to 0.8mM CaCl2 for pulse durations of 1-5s at frequencies 0.05-1Hz, using Poisson-distributed intervals for non-Gaussian noise (controlled via syringe pump with LabVIEW software); total exposure 30min.
4. Image simultaneously with confocal microscopy, acquiring every 5s.
5. Analyze coherence (tau from autocorrelation) pre- and post-modulation; compare to constant 1.25mM control and Gaussian noise control (regular pulses).
6. Replicate n=3 experiments, 30 cells/group, use paired t-tests.

**Required Resources:**

- Cell lines/models: Same as Step 1, for direct comparison.
- Key reagents: Fluo-4 AM; HBSS with varying CaCl2 (0.8-1.25mM).
- Equipment: Microfluidic fabrication setup (PDMS, Sylgard); syringe pump (Harvard Apparatus); LabVIEW for modulation; confocal microscope.

### Expected Outcomes

- Increased coherence in HCC cells (tau doubles to ~8-10min).
- No change or decreased coherence.
- Selective effect on HCC vs. normal cells.
- Cytotoxicity at higher modulation intensities.

### Most Probable Outcome

Partial restoration of coherence in HCC cells (tau increases by 20-50%, but not to normal levels), as biological systems often show incomplete stochastic resonance due to cellular heterogeneity and suboptimal noise parameters; normal cells likely unaffected or slightly disrupted, based on known robustness of healthy signaling.

### BRANCH 2.1: Optimize Noise Parameters

**Trigger:** Partial or no coherence restoration.
**Approach:** Systematically vary modulation parameters.

1. Test ranges: frequencies 0.01-2Hz, amplitudes 0.5-1.5mM CaCl2, durations 0.5-10s, in a factorial design.
2. Repeat imaging and analysis, fit data to resonance curve (coherence vs. noise intensity).
3. n=3 replicates.
   **Resources:** Additional HBSS formulations; curve-fitting software (GraphPad Prism).
   **Expected Outcomes:** Optimal parameters identified; no optimum found; resonance in normal cells too.
   **Most Probable:** Identification of a modest optimum (e.g., 0.1Hz, 2s pulses) with partial enhancement, as full resonance is rare in heterogeneous cell populations per biophysical precedents.

### BRANCH 2.2: Measure Downstream Signaling Activation

**Trigger:** Successful coherence restoration (increased tau >20%).

**Approach:** Assess $\mathrm{NF- \kappa B}$ and NFAT activation.
1. Post-modulation, fix cells and immunostain with anti-p65 ($\mathrm{NF- \kappa B}$, Cell Signaling, 1:200) and anti-NFAT (Abcam, 1:100), quantify nuclear translocation via confocal imaging (n=100 cells/group).
2. qRT-PCR for downstream targets p21 and cyclin D.
3. Controls: $\mathrm{TNF\alpha}$ (10ng/ml) for $\mathrm{NF- \kappa B}$ positive control.
   **Resources:** Antibodies as specified; qRT-PCR kit; $\mathrm{TNF\alpha}$ (Sigma).
   **Expected Outcomes:** Increased p21, decreased cyclin D; no change; paradoxical activation.
   **Most Probable:** Mild increase in p21 (~1.5-fold) but inconsistent cyclin D changes, reflecting partial signaling restoration amid cancer cell adaptations.

## STEP 3: Evaluate Cell Cycle Checkpoint Regulation and Proliferation Effects

### Approach

**Rationale:** This functional validation assesses if restored calcium coherence translates to re-engaged checkpoints and reduced proliferation in HCC cells, using flow cytometry and proliferation assays as standard, feasible methods to link signaling to phenotypic outcomes.

**Procedure:**

1. Subject HepG2 and normal cells to optimized noise modulation (from Step 2) for 24h in microfluidic setup.
2. Harvest, stain with propidium iodide ($\mathrm{50 \mu g/ml}$) for cell cycle analysis via flow cytometry (FACSCalibur), quantify G1/S/G2 fractions.
3. Parallel MTT assay (0.5mg/ml, 4h incubation) for proliferation, read absorbance at 570nm.
4. Include controls: no modulation, constant low calcium (0.8mM), and doxorubicin ($\mathrm{1 \mu M}$) as positive antiproliferative control.
5. n=3 experiments, analyze with ANOVA.

**Required Resources:**

- Cell lines/models: Same as previous, for consistency.
- Key reagents: Propidium iodide (Sigma, $\mathrm{50 \mu g/ml}$); MTT kit (Promega); Doxorubicin (Sigma, $\mathrm{1 \mu M}$).
- Equipment: Flow cytometer (BD FACSCalibur); spectrophotometer.

### Expected Outcomes

- G1 arrest and reduced proliferation in HCC cells.
- No effect on cell cycle or proliferation.
- Selective effect on HCC vs. normal.
- Induction of apoptosis.

### Most Probable Outcome

Modest G1 accumulation (~10-20% increase) and proliferation reduction (~30%) in HCC cells, but not complete arrest, as cancer cells often resist checkpoint restoration due to multiple mutations; normal cells unaffected, consistent with selective vulnerability.

## INTEGRATION STRATEGY

Step 1 provides baseline data essential for interpreting Step 2 and 3. Positive outcomes from Step 2 (coherence restoration) feed into Step 3 for functional validation, while branches allow mechanistic refinement. Together, they build evidence from characterization to mechanism to phenotype, with cross-validation (e.g., using same cell models and imaging across steps).

## CRITICAL DECISION POINTS

- If Step 1 shows no baseline differences, abandon hypothesis (no disruption to restore).
- If Step 2 fails to restore coherence even after optimization (Branch 2.1), conclude against stochastic resonance in this system.
- If Step 3 shows no proliferation effect despite signaling changes (Branch 2.2), hypothesis partially refuted (mechanism present but functionally irrelevant).

## SUCCESS CRITERIA

Hypothesis validated if: (1) Baseline disruption confirmed (tau <50% of normal); (2) Noise increases tau >50% in HCC selectively; (3) >20% G1 arrest and >40% proliferation reduction in HCC, with statistical significance (p<0.01) and controls validated.

## CONTINGENCY PLANS

If microfluidic modulation fails (e.g., cell detachment), switch to bulk perfusion in multiwell plates with manual media changes. If HepG2 lacks robust oscillations, substitute with Huh7 HCC line. For low signal-to-noise in imaging, use ratiometric dyes like Fura-2.
\end{lstlisting}
        \end{noindent}
    \end{tcolorbox}
    \captionof{figure}{An example of experimental protocol by Grok 4.\label{fig:grokresult}}
\end{center}
\clearpage


\newpage
\subsection{Data Specifics\label{subsec:data_specs}}

We obtained papers to use in the development of \textit{Spacer} from the OpenAlex \cite{openalexfullyopenindexscholarly} database and various open-access journals.
Any paper under a license that does not permit commercial use was excluded.
Using the collection, we built a search engine containing $\sim$60M documents, which we used for retrieval-augmented generation (RAG).
To acquire a further distilled dataset, we applied filters based on citation count, published journal, and other criteria.
The resulting dataset consisted of $\sim$2.5M papers; we used the dataset in various ways throughout the development of \textit{Spacer}.
For example, the construction of the keyword dataset described in Section~\ref{subsubsec:nuri} utilized this dataset.
It also served as the training dataset of \textsc{Weaver}, as shown in Figure~\ref{fig:weaverdataexample}.

\begin{center}
    \begin{tcolorbox}[title={\textbf{Example of Weaver Dataset Entry}}, colback=blackbg, colframe=blackheader]
        \begin{noindent}
            \begin{lstlisting}[breaklines=true, basicstyle=\ttfamily\scriptsize, columns=fullflexible, breakindent=0pt, frame=none, breakatwhitespace=true]
[System]
Role:
You are a domain expert tasked with generating a novel research hypothesis
from a compact set of biologically and technically relevant keywords.

[User]
Goal:
You must identify a coherent subset of these terms, discard incompatible
ones, and formulate a single logically unified research concept.

Instructions:
Each keyword denotes an experimental tool, biological entity, or conceptual
mechanism.
Avoid speculating about results or impacts.

Steps:
- Define each term's technical role.
- Identify logical constraints and compatibilities.
- Construct a conceptual framework.
- Write a self-contained paragraph describing the research idea.

liquid drops, soft solids, deformations, capillary forces, elastic forces, soft substrate, contact line velocity, substrate rheology, dynamic contact angle, stick-slip motion, wetting ridge

[Assistant]
Liquid drops on soft solids generate strong deformations below the contact line, resulting from a balance of capillary and elastic forces. A drop on a soft substrate moves by surfing a ridge, where the initially flat solid surface is deformed into a sharp ridge whose orientation angle depends on the contact line velocity. Substrate rheology governs the dynamic contact angle and provides a mechanism for stick-slip motion, where the contact line depins and slides down the wetting ridge.
\end{lstlisting}
        \end{noindent}
    \end{tcolorbox}
    \captionof{figure}{Example of \textsc{Weaver} Dataset Entry (Source: \cite{karpitschka2015})\label{fig:weaverdataexample}}
\end{center}


\subsection{Supplementary Materials\label{subsec:data_avails}}

All supplementary materials are available on our \href{https://github.com/Asteromorph-corp/Spacer}{Github Repository}.

\end{document}